\documentclass{article}
\usepackage[english]{babel}
\usepackage[letterpaper,top=2cm,bottom=2cm,left=3cm,right=3cm,marginparwidth=1.75cm]{geometry}
\usepackage{amssymb}
\usepackage{amsmath}
\usepackage{graphicx}
\usepackage[colorlinks=true, allcolors=blue]{hyperref}

\newtheorem{theorem}{Theorem}

\newtheorem{definition}{Definition}
\usepackage{algorithm2e}

\usepackage{dirtytalk}
\usepackage{csquotes}

\usepackage{mathtools}

\DeclarePairedDelimiter\floor{\lfloor}{\rfloor}

\usepackage{setspace}
\title{Approximation Methods for Partially Observed Markov Decision Processes (POMDPs)}
\author{Caleb M. Bowyer}

\begin{document}

\maketitle
\begin{abstract}

POMDPs are useful models for systems where the true underlying state is not known completely to an outside observer; the outside observer incompletely knows the true state of the system, and observes a noisy version of the true system state. When the number of system states is large in a POMDP that often necessitates the use of approximation methods to obtain near optimal solutions for control. This survey is centered around the origins, theory, and approximations of finite-state POMDPs. In order to understand POMDPs, it is required to have an understanding of finite-state Markov Decision Processes (MDPs) in \autoref{mdp} and Hidden Markov Models (HMMs) in \autoref{hmm}. For this background theory, I provide only essential details on MDPs and HMMs and leave longer expositions to textbook treatments before diving into the main topics of POMDPs. Once the required background is covered, the POMDP is introduced in \autoref{pomdp}. The origins of the POMDP are explained in the classical papers section \autoref{classical}. Once the high computational requirements are understood from the exact methodological point of view, the main approximation methods are surveyed in \autoref{approximations}.  Then, I end the survey with some new research directions in \autoref{conclusion}



\end{abstract}

\tableofcontents

\section{A Brief Review of MDPs}
\label{mdp}
This section discusses background material on MDPs using several popular textbooks on the subject. Two modern textbooks I recommend for MDPs are \cite{bertsekasdynamic1} and \cite{bertsekasdynamic2}. Reading selective parts of \cite{bertsekasdynamic1}, updated in 2017, is sufficient to continue towards an introductory treatment of POMDPs. One other textbook I recommend that adds perspective to this text, and has additional material for MDPs as well as an excellent coverage of POMDP theory is the book \cite{krishnamurthy2016partially}. Published in 2016, the text is a modern treatment of POMDPs with many examples of POMDPs and an easy to read development of POMDP theory. Only the essential parts of MDPs and HMMs are covered in this introductory chapter. Moreover, I provide a quick survey of references to some other areas of interest to MDPs, mostly extensions to the basic theory that has been achieved that may have an impact on future research for POMDPs. 






\subsection{The Dynamic Programming (DP) Algorithm}
DP to MDP reading Road map in \cite{bertsekasdynamic1}:\newline
\begin{enumerate}
    \item Read first appendix D On Finite-State Markov Chains if no background with Markov chains is had. Then, proceed to Chapter 1.
    \item Read Chapter 1 The DP Algorithm.
    \item Other mathematical background required and assumed is basic familiarity with calculus, linear algebra, probability and Markov chains. \item Within this first chapter of \cite{bertsekasdynamic1} particular emphasis should be placed on 1.1.2 Discrete-State and Finite-State Problems p.7 as this is the setting that leads towards a development of HMMs. Then, one can approach a suitable POMDP introductory development for finite-state based systems. Note that Bertsekas's coverage includes continuous valued state-spaces, and allows for easy addition of disturbances to the state from a countable set in his coverage of the dynamic programming algorithm. 
    \item Next, read 1.2 The basic problem, 1.3 The Dynamic Programming Algorithm. Skip section 1.4 State Augmentation and Other Reformulations. Read section 1.5 Some Mathematical issues. Optional: read 1.6 Dynamic Programming and Minimax Control.
\end{enumerate}

Summary:
this chapter helps one come to understand the general structure of the DP algorithm for a finite horizon optimal control problem i.e., over a finite number of $N$ stages. The two main features of the problem are a discrete-time dynamic system that evolves the state, and an additive cost function. Notation for common variables is introduced: $x_k$ is the state at time $k$, $f_k$ is the system function at time $k$ that takes in parameters $(x_k, u_k, w_k).$ The disturbance (noise) $w_k$ comes from a finite set, viewed as a random parameter. The discrete time index is $k \in \{0, 1,\ldots, N-1\}$. The length of the horizon is $N$ and is the number of times control is applied by an agent. The control or decision an agent selects is denoted as $u_k$ at time $k$. The total cost for this problem is $c_N(x_N) + \sum\limits_{k=0}^{N-1} c_k(x_k, u_k, w_k)$. Here the terminal cost is $c_N(x_N)$. The basic problem is formulated in terms of minimizing the expected sum of these costs through an optimal policy (how an optimal controller decides on a control when in some state at some time).

In \cite{bertsekasdynamic1}, an important distinction between open loop vs. closed-loop minimization is made. In open loop minimization, one selects $u_0,\ldots,u_{N-1}$ at the initial time $k=0$ without considering what the future states and disturbances actually are. On the other hand, in closed-loop control an agents waits until the current state $x_k$ is known at stage $k$ before reaching a control decision. The goal of DP and MDPs specifically is to find an optimal rule for selecting $u_k$ for any state $x_k$ in the state space. One wants the sequence of functions $\mu_k$ for $k \in \{0,1,\ldots,N-1\}$ mapping $x_k \rightarrow u_k$ to minimize the expected cumulative cost. For the finite horizon there are most commonly $N$ distinct policies $\mu_k$, whereas for the infinite horizon, the policy is stationary and denoted as $\mu$, without the time index $k$. The sequence $\pi = \{\mu_0,...,\mu_{N-1}\}$ is called the policy or control law for the finite horizon. For each policy $\pi$, the expected cost of $\pi$ starting from state $x_0$ is $$J_\pi(x_0) = E\biggl\{\sum\limits_{k=0}^{N-1} c_k(x_k, \mu_k(x_k), w_k) + c_N(x_N)\biggr\}.$$ 


\subsection{MDP Basics: The Finite Horizon Case}
Also, consider reading \cite{krishnamurthy2016partially} Ch. $6$ Fully Observed Markov Decision Processes. In particular, read Ch $6.1$ Finite state finite horizon MDP, and $6.2$ Bellman's stochastic dynamic programming algorithm. An MDP is a Markov chain with a control applied at every stage. At some stage $k$, a state $x_k$ of the world is occupied by an agent or decision maker (controller) where it chooses a control $u_k$; this choice selects a transition matrix $P(u_k)$, which is of size $n\times n$ for an $n$-state Markov chain. The choice of control input $u_k$ influences the probability that the next state is some value $X_{k+1}$ at time $k+1$. The goal is to choose controls over the horizon, infinite or finite in stage number, that minimize some notion of cost termed a cost-to-go function. Most of the time, these cost-to-go functions are cumulative or additive of the costs received over the stages of the decision making. 

First, it should be pointed out that the finite horizon MDP which is N stages long is fully specified by a $5$ tuple: $(\mathcal{X},\mathcal{U}, P_{ij}(u), c(i,u), c_N(i))$ and possibly a discount factor $\gamma$. For every stage in the horizon, an important point for MDPs is that the agent knows the model parameters specified by the $5$ tuple and that there is no corruption or delay in the state information to the agent or controller $\forall k \in \{0,1,\ldots,N\}$. The finite state space is denoted by $\mathcal{X} = \{1,2,\ldots,n\}$, and the finite control space is denoted by $\mathcal{U} = \{1,2,\ldots,m\}$. Additionally, the probability of transitioning from state $i$ to state $j$ at time $k$ when control $u$ is applied is denoted by 

$$P_{ij}(u) = P(X_{k+1}=j|X_k=i,U_k=u)~\forall i,j\in \mathcal{X},~u\in \mathcal{U},~k\in \{0,1,\ldots,N-1\}.$$

Furthermore, $c(i,u)$ denotes the instantaneous cost of being in state $i$ and executing control $u$. Lastly, $c_N(i)$ denotes the terminal cost $\forall i \in \mathcal{X}$. An important fact of MDPs in terms of the policy is that only the current state $x_k$ needs to be considered for making an optimal decision, not the full information set $\mathcal{I}_k = \{x_0,u_0,\ldots,u_{k-1},x_k\}$. In general, $u_k = \mu_k(\mathcal{I}_k)$ maps from the information set to the control space $\mathcal{U}$; a policy, also called a decision rule for MDPs is a mapping from $\mathcal{X}$ to $\mathcal{U}$, i.e., $u_k=\mu_k(x_k)$ denotes a deterministic mapping from $\mathcal{X}$ to $\mathcal{U}$. 

Another change in notation from \cite{bertsekasdynamic1} to \cite{krishnamurthy2016partially} is that $\underbar{$\mu$}$ is a policy whereas $\pi$ denotes a belief state when this is encountered in the POMDP chapters. The notations $b$ and $\pi$ are both common for belief-states. The preceding type of policy is called a deterministic Markovian policy and this is sufficient to achieve the minimum of:

\begin{equation*}
\begin{split}
J_{\underbar{$\mu$}}(x) &= E_{\underbar{$\mu$}}\{\sum_{k=0}^{N-1} c(x_k,\mu_k(\mathcal{I}_k)) + c_N(x_N)|X_0 = x\} \\
                      &= E_{\underbar{$\mu$}}\{\sum_{k=0}^{N-1} c(x_k,\mu_k(x_k)) + c_N(x_N)|X_0 = x\},
\end{split}
\end{equation*}
which is the expected cumulative cost of the policy $\underbar{$\mu$} = (\mu_0, \mu_1,\ldots, \mu_{N-1})$ over the finite length $N$ horizon. The goal of the agent or decision maker is to determine $\underbar{$\mu$}^* = \arg \min \limits_{\underbar{$\mu$}} J_{\underbar{$\mu$}}(x)$, i.e., find a policy that minimizes the expected cumulative cost for all initial states $x_0$. When $\mathcal{X}$ and $\mathcal{U}$ are finite spaces there always exist an optimal policy, however uniqueness is not guaranteed. The solution to this problem of finding the optimal policy comes from Bellman's stochastic dynamic programming algorithm. The optimal policy $\underbar{$\mu$}^* = (\mu_0, \mu_1,\ldots,\mu_{N-1})$ is obtained from a backward recursion. See p.$128$ of \cite{krishnamurthy2016partially} for the details. After running this algorithm for $N$ steps, the optimal expected cumulative cost-to-go is given by $J_0(i)$ for all starting states $i$ from this algorithm. Furthermore, at each stage Bellman's algorithm constructs the optimal deterministic policy $\mu_k^*(i)$ for all possible $i\in \mathcal{X}$. The proof of this algorithm relies on the principle of optimality. See p.$125$ of \cite{krishnamurthy2016partially} and page $20$ of \cite{bertsekasdynamic1}. In short, given any optimal policy for an arbitrary horizon, if the problem is started at an arbitrary stage in an arbitrary state, the original optimal policy for the full horizon is still optimal for the tail or sub-problem whenever and wherever the agent starts executing the optimal policy.

\subsection{MDP Basics: The Infinite Horizon Case}

$~~~~$ The only other material I would like to cover in this MDP section is the extension to the infinite horizon case, including both discounted total cost with its several cases and a discussion of the average cost case for the different possible cost-to-go function formulations. The famous computational methods are given as assigned reading, as part of the reading road-map.

In Ch. $5$ of \cite{bertsekasdynamic1}, Introduction to Infinite Horizon Problems, this chapter should be read in its entirety except the Semi-Markov Problems section $5.6$ can be safely skipped. All other sections should be read multiple times to let the details sink in to the mind because some of the facts are quite subtle. From an MDP perspective, the infinite time horizon is a natural extension of the finite time horizon. There are several nice features of this chapter worth reading, particularly the connection between stochastic shortest path problems and DP, the practical computational algorithms, and the kinds of cost-to-go functions that can be optimized. Read 5.1 an overview, 5.2 Stochastic Shortest Path Problems, 5.3 Computational Methods, 5.4 Discounted Problems, 5.5 Average Cost per Stage Problems.

Infinite horizon summary: extra care must be taken in the analysis of these DP and MDP equations as the horizon length goes to infinity in any limiting arguments made for proofs. The model differs from the previous section in the following ways:
\begin{enumerate}
    \item The number of stages is now countably infinite.
    \item The system and cost functions are stationary.
    \item The policy is also stationary.
\end{enumerate}
There are four general types of infinite horizon problems discussed 1) minimization of total discounted cost:

$$ J_\pi(x_0) = \lim_{N\to\infty} \displaystyle E\biggl\{\sum_{k=1}^{N-1}\alpha^k g(x_k,\mu_k(x_k),w_k)\biggr\},$$
where the expectation is taken with respect to the $w_k$. Here $\pi=(\mu_0,\mu_1,\ldots)$ and $0<=\alpha<=1$ is called the discount factor for the problem. The meaning of $\alpha<1$ is that future costs of the same magnitude matter less in the future than they currently do at the present time. Something that is often a concern is whether the limiting cost-to-go, $J_\pi(x_0)$, is finite. This can be ensured through sufficient discounting.


There are a few different cases to consider for the infinite horizon definition from \cite{bertsekasdynamic1}:
\begin{displayquote}
\begin{enumerate}
    \item Case a) stochastic shortest path problems. Here, $\alpha=1$, but there must be at least one cost-free termination state; Once the system reaches one of these states, it remains there at no further cost. I have also heard this state referred to as a graveyard state before. In effect, this is a finite horizon problem of a random horizon length, and it is affected by the policy being used. For instance, a policy could be prohibiting an agent from reaching a terminal state faster by making poor choices, thus incurring more costs than is necessary.
    \item Case b) discounted problems with bounded cost per stage:
  Here, $\alpha < 1$ and the absolute cost per stage $|g(x,u,w)|$ is bounded from above by some constant M; this makes the limit in the definition of the cost $J_\pi(x_0)$ well defined because it is the infinite sum of a sequence of numbers that are bounded in absolute value by the decreasing geometric progression $\{\alpha^kM\}$.
  \item For Case c) Discounted and Undiscounted problems with unbounded cost per stage: This analysis is not done in \cite{bertsekasdynamic1}. See Chapter 4 of \cite{bertsekasdynamic2} for further details. The main issue is the handling of infinite costs that some policies achieve.
  \item The last case, Case d) Average cost per stage problems is dealt with in Ch. 5 of \cite{bertsekasdynamic1}.
\end{enumerate} 
\end{displayquote}

The average cost-to-go formulation makes sense when $J_\pi(x_0)=\infty$ for every policy $\pi$ and initial state $x_0$. This explosion in the cost-to-go could happen, for instance, if $\alpha=1$ and the cost $g$ per stage is positive. Then, it makes sense to turn to a notion of average cost per stage, to properly study the infinite horizon. The average cost-per-stage is defined as $$ \lim_{N\to\infty} \frac{1}{N} E\biggl\{\sum_{k=1}^{N-1}\alpha^k g(x_k,\mu_k(x_k),w_k)\biggr\},$$ and is often well defined, i.e., finite.

  
Now that these types of the infinite horizon problem are presented, I want to show an interesting fact that is valuable both analytically, and computationally. Taking the limit in $N$ of the finite horizon cost-to-go function yields the infinite horizon cost-to-go, i.e., $J^*(x) = \lim\limits_{N\to\infty}J_N(x)$. This relationship typically holds; it holds for Cases a-b), but exceptions exist for Case c). The following equation is Bellman's equation or Bellman's optimality equation, for the stochastic control setting:

  $$ J^*(x) = \displaystyle\min_{u\in U(x)}\displaystyle E\biggl[g(x,u,w) + \alpha J^*(f(x,u,w))\biggr].$$
  
This is a system of $n$ equations. The solution to this set of equations is the optimal cost-to-go function. It is typically hard to compute the optimal cost-to-go exactly, but algorithms exist that converge to the correct solution. Most algorithms rely on the use of a mathematical property called contraction. The notion of a contraction mapping from one cost-to-go function into another is more mathematically advanced and the viewpoint is not needed for the purposes of this survey paper. More pointers are contained in \cite{bertsekasdynamic2}. There one can read about all of the computational methods, especially value iteration, and policy iteration as they will have parallels to algorithms to be developed for POMDPs. 




Additional reading for this section can be found in Ch.6 of \cite{krishnamurthy2016partially} to deepen understanding of MDPs and gain more perspective. Specifically, compare the MDP development of the infinite horizon of section 6.4 with the infinite horizon discounted cost to the form of the above DP Bellman equation. Remember, the infinite-horizon model consists of a 5-tuple $(\mathcal{X},\mathcal{U}, P_{ij}(u), c(i,u), \alpha)~\forall i,j\in \mathcal{X}, u\in\mathcal{U}$. Because the horizon is infinite, note that the terminal cost has been removed. The limiting form of the finite horizon MDP Bellman equation is: $$ J^*(i) = \displaystyle\min_{u\in U(x)} \biggl\{c(i,u) + \alpha \sum_{j=1}^{n}P_{ij}(u) J^*(j) \biggr\} .$$ Note that the expectation is removed for this equation, and the dynamics that was in the system equation $f$ has moved to the matrix $P$ which acts like a conditional expectation operator. Rather than include an explicit random variable $w$ in the system of equations, the randomness comes from $P$ as does the dynamics.
\section{The Hidden Markov Model (HMM)}
\label{hmm}

\subsection{HMM Basics and the Viterbi Algorithm}
Throughout this survey paper I use unified notation when I review all the papers and update the original paper's notation to match mine where appropriate so the reader is not stuck deciphering different notations and conventions. After following the above reading recommendations for MDPs, this section continues with the introduction of HMMs. One could continue reading \cite{bertsekasdynamic1} for HMM material or go to \cite{krishnamurthy2016partially} and begin reading on stochastic models and Bayesian filtering. The Bayesian filtering perspective is crucial for appreciating the HMM in the eventual POMDP context. The filter is used to update the agent or controller's belief-state as new information, specifically observations of the Markov chain as they are witnessed. This belief-state formulation is important because most of the time the agent has inaccurate and corrupt state information to determine its control decisions.  In this context, it is important to determine the true state of the system as best as possible. In this section I review only the most useful information related to HMMs from \cite{bertsekasdynamic1} and \cite{krishnamurthy2016partially} to provide understanding of the HMMs before in the next section moving on to POMDPs, the main focus of the survey.

Reading road-map:
\begin{enumerate}
    \item In \cite{bertsekasdynamic1} Chapter 2, read Deterministic Systems and the Shortest Path Problems

    \item Read 2.1 Finite-State Systems and Shortest Paths. This section provides an interesting perspective.
    
    \item Then, jump to the section 2.2.2 on Hidden Markov Models and the Viterbi Algorithm. 
\end{enumerate}
Later, one will see how the HMM is a critical part of POMDPs. Section 2.2.1 can be skipped. And skip the rest of this chapter, mainly focused on shortest path algorithms for graphs. 

Summary of HMMs from \cite{bertsekasdynamic1}: A Markov chain is considered, which has a finite state space with given transition probabilities $P_{ij}$. The point of the HMM is to optimally estimate the unknown states as the Markov chain evolves over time. Assume an observer can only see observations that are probabilistically related to the true state. The observer can optimally estimate the state sequence from the observation sequence. 
Denote the probability of an observation $z$, after a state transition from $i$ to $j$ by $r(z;i,j)$. The observer is also given the prior $\pi_i$ on initial state probabilities, i.e. the probability the state initially takes on some value $i$. The probablistic transitions just describe constitutes an HMM, which is a noisly observed Markov chain. Note for future reference that when control is added to the model, a POMDP is formed. The Viterbi algorithm uses the \say{most likely state} estimation criterion such that it is able to take the observation sequence $Z_n = \{z_1, z_2, \ldots, z_N\}$ and compute the estimated state transition sequence $\hat{X}_n = \{\hat{x}_0, \hat{x}_1,\ldots,\hat{x}_N\}.$

In \cite{krishnamurthy2016partially} read Ch.2 Stochastic state space models, and specifically read 2.1 Stochastic state space model, 2.2 Optimal prediction: Chapman-Kolmogorov equation, and 2.4 Example 2: Finite-state hidden Markov model (HMM). Then, proceed to Ch. 3 Optimal filtering. In this chapter, read 3.1 Optimal state estimation, 3.2 Conditional expectation and Bregman loss functions, 3.3 Optimal filtering, prediction and smoothing formulas. Section 3.3 is important for understanding optimal filtering in the most general space setting. Then, immediately read 3.5 Hidden Markov model (HMM) specializing the previous results to finite-state Markov chains. Lastly, in this chapter, read 3.7, an important concept of Geometric ergodicity of HMM filter which says that an HMM \say{forgets} its initial condition geometrically fast, like a Markov chain does under certain conditions.

\subsection{Optimal Prediction: Chapman-Kolmogorov (C-K) equation}

Summary from \cite{krishnamurthy2016partially}: an interesting question for HMMs is what is the optimal state predictor? For a finite-state Markov chain with state-space $\mathcal{X}=\{1,2,\ldots,n\}$ and a state transition matrix $P$, define the $n$-dimensional state probability vector at time $k$ as $$\pi_k = [Pr(X_k=1),\ldots,Pr(X_k=n)]^T.$$ Then, the Chapman-Kolmogorov (C-K) equation reads:
$$ \pi_{k+1} = P^T \pi_k, $$ initialised by some prior belief $\pi_0$. The process $\{X_k\}$ denotes an $n$-state Markov chain. The dynamics of $X_k$ are described by the $n\times n$ $P$ matrix with elements $P_{ij}~\forall\  i,j \in \{1,2,\ldots,n\}\  s.t.\  0\leq P_{ij} \leq 1,\  \sum\limits_{j=1}^{n}P_{ij} = 1 \forall i$. Associated with the states $1,2,\ldots,n$, one can define a state level vector $C = [C(1), C(2),\ldots,C(n)]^T$. Using $\pi_{k+1}$ we can predict the state level at time $k+1$ by $E[C(X_{k+1})] = \sum\limits_{i=1}^{n}C(i)\pi_{k+1}(i)$. 

A few more definitions of Markov chains may prove useful for proofs or analysis, namely the limiting distribution and the stationary distribution. The limiting distribution: $ \lim\limits_{k\to\infty} \pi_k = \lim\limits_{k\to\infty}(P^T)^k\pi_0$, which may not exists as a limit, but follows from repeated application of the C-K equation. Another important definition for Markov chains is the Stationary distribution which is the $n$-dimensional vector $\pi_{\infty}$ that satisfies $\pi_{\infty} = P^T\pi_{\infty},~\bigl < \mathbf{1},\pi_{\infty}\bigr > = 1$. The vector $\pi_{\infty}$ is interpreted as a normalized right eigenvector of $P^T$ corresponding to the unit eigenvalue. The limiting distribution and the stationary distribution coincide if $P$ is regular. The definition of regular is: A transition matrix $P$ is regular (primitive) if $\exists k>0$ s.t.$P^k_{ij}>0\  \forall i,j \in \mathcal{X}$.

\subsection{Optimal Filtering for the HMM}

Now, I present the main results from Ch.3 on Optimal filtering from \cite{krishnamurthy2016partially} necessary for our purposes of building to the POMDP theory. The recursive state estimation problem goes by several names: optimal filtering, Bayesian filtering, stochastic filtering, recursive Bayesian estimation, and so on. The name is not so important, but what is important is that optimal filtering involves the use of Bayes' rule.

Next, I present the HMM filter. As before, one has the state transition probabilities (uncontrolled) $P_{ij} = Pr(X_{k+1}=j|X_k=i)$, but additionally one has another finite space for the observations $\mathcal{Y} = \{1,2,\ldots,Y\}$. It is also possible to have continuous values for the case of noisy observed values of the HMM. See section 2.4 of \cite{krishnamurthy2016partially} for more on this. Effectively, one has to use probability distributions to model that effect.  For the HMM setting, the conditional probability mass function (pmf) is defined as 
$$B_{xy} = Pr(Y_k=y|X_k=x)\  \forall x\in \mathcal{X}, y\in \mathcal{Y}.$$ 
These jumps are also called state-to-observation transitions. Now, we have state-to-state transitions, and state-to-observation transitions as part of our modeling procedures. Most applications of HMMs such as speech processing or telecommunications use a finite observation space. $B$ can be thought of as a $X\times Y$ matrix that satisfies $\sum\limits_{y=1}^{Y}B_{xy} = 1$.

Let's recap the HMM model before discussing the optimal filtering update to the belief state. An HMM is a noisily observed Markov chain specified by a 4-tuple $(P, B, C, \pi_0)$, i.e., the state and observation transition probabilities, the state level vector, and an initial belief pmf. When dealing with continuous random variables, the update formula is given by integrals, but for finite state spaces, the integration reduces to summation:
$$ \frac{Pr(Y_{k+1}|X_{k+1}=j)\sum\limits_{i=1}^{n}P_{ij}\pi_k(i)}{\sum\limits_{l=1}^{n}Pr(Y_{k+1}|X_{k+1}=l) \sum\limits_{i=1}^{n}P_{il}\pi_k(i)}.$$
This equation can be expressed beautifully and compactly using matrix-vector notation. Having observed the value $Y_k$, define the $n\times n$ diagonal matrix of likelihoods:
$$ B_{y_k} = \textit{diag}(p(y_k|x_k=1),\ldots,p(y_k|x_k=X)).$$ Define the $n$-dimensional posterior vector, also called the \say{belief-state} as $\pi_k = [\pi_k(1),\ldots,\pi_k(n)]^T$. For observations $y_k,~k=1,2,\ldots$, the HMM filter is summarized now. Given known parameters $(P, B, C, \pi_0)$, for time $k=0,1,\ldots$, given observation $y_{k+1}$, update the belief-state by $$ \pi_{k+1} = T(\pi_k, y_{k+1}) = \frac{B_{y_{k+1}}P^T\pi_k}{\sigma(\pi_k, y_{k+1})}, $$ where $\sigma(\pi_k, y_{k+1})= \mathbf{1}^T B_{y_{k+1}}P^T\pi_k$ is our normalizing constant.

Next, I present a few different types of state estimators for the HMM that could be useful for a number of reasons once we reach POMDP theory in the next chapter. First, define a type I estimator at time $k$ as $$ \hat{X}^{(1)}_k = \arg\max_{1\leq i \leq X} \pi_k(i). $$ This type of estimation could be useful for application in POMDPs as discussed shortly. Secondly, I introduce another type of state estimator that could be useful when the number of states is very large even though still finite in number. Denote as the type II state estimator, and when the state levels take on the state values reduces to the following form:
\begin{equation*}
\begin{split}
\hat{X}^{(2)}_k &= \textit{NN}(\floor{E\{C(X_k)\}})\\
                &= \textit{NN}(\floor{E\{X_k\}})\\
                &= \textit{NN}(\floor{\sum_{i=1}^{X}i\pi_k}),
\end{split}
\end{equation*}
where NN stands for nearest neighbor here. Certainly other forms of state estimators exist and should be experimented with compared to these. Those presented now are a few possibilities.

After having developed some HMM theory, I have presented two simple state estimators, called type I and type II that could be used in the previously developed DP and MDP material to solve the HMM and later in POMDPs by turning it into an MDP. More importantly, it provides us approximations or efficient computations of the kind discussed above for solving the HMM, although certainly other variations on these ideas are possible. Later, more types of state estimators for approximations to the POMDP problem are realized after surveying the classical papers. Before reviewing in detail the classical papers, the POMDP is motivated and introduced in the next section. 
\section{POMDP Fundamentals}
\label{pomdp}

$~~~~$POMDPs generalize MDPs when state uncertainty exists or partial information is available to an agent to make decisions. The POMDP captures both stochastic state transitions and observation uncertainty. The canonical model includes finite state, finite control, and finite observation spaces. This model is very general and has been applied to solve research problems in a number of fields. A survey on POMDP applications is found in \cite{cassandra1998survey}.

\label{Canonical_Model}

In MDPs, the state $x_k$ is observable to an agent at all stages. Meanwhile, in a POMDP, the agent can only observe an output or observation $y_k$, which can in many cases be treated as a noisy version of $x_k$.

\begin{definition}
The POMDP Canonical Model consists of:
\begin{enumerate}
  \item A finite set of states $\mathcal{X}=\{1,2,\ldots,n\}$.
  \item A finite set of controls $\mathcal{U}=\{1,2,\ldots,l\}$.
  \item A finite set of observations $\mathcal{Y}=\{1,2,\ldots,m\}$.
  \item A state-to-state transition function $p_{ij}(u) = Pr(X_{k+1}=j|X_k=i,U_k=u)$.
  \item A state-to-observation transition function $q_{jy}(u) = Pr(Y_{k+1}=y|X_{k+1}=j,U_k=u)$.
  \item A cost function, most generically defined on $\mathcal{X} \times \mathcal{U} \times \mathcal{Y}$ as $c(i,u,y)$.
\end{enumerate}
\end{definition}
The immediate cost $c(i,u,y)$ is the most generic kind of cost function for this model because it considers not only the current state $x$ and control $u$ executed from $x$, but also any observation $y$ that can be observed. If observations are not relevant for the cost model, then the immediate cost can be written as $c(i,u)$. This model can be used to learn a control policy that yields the lowest total cost over all stages of decision making. Also note that the above sets could be modeled as continuous with some more research effort. In fact, in my most recent unpublished research \cite{bowyer2021pomdp}, the observation space was modelled as being continuous, but the state and control sets were still modelled as being finite. More research is needed on allowing for all of the spaces to be continuous and being able to determine optimal or near-optimal controllers for such models. 




\label{Exact_Methods}
A dissertation by Alvin Drake at MIT \cite{drake1962observation}, was the first work akin to POMDPs that studied decoding outputs from a noisy channel. He was the first person to study what we now call POMDPs by looking at decoding Markov sources over a noisy channel. In his work, the states are the symbols of the Markov source and the observations are the symbols received in noise. He considered a binary symmetric channel model of the noise with some different decoding rules. The extension to continuous output alphabets, the inclusion of error control coding, and the potential application of RL to such problems remain open research problems. 

In an earlier paper \cite{bowyer2019reinforcement}, I investigated RL for solving the spectrum sharing problem, using the simple SARSA algorithm as the main learning technique. Better modelling of all of the agents would occur if a POMDP was able to be formulated; This paper, \cite{wong2020dynamic}, explains the radio architecture and more modelling considerations for the RL radio problem. Additionally, a better multi-agent formulation of the multi-radio spectrum sharing problem would bring about more performance gains in terms of optimally dividing the spectrum between all teams' radios. Note that the general multi-radio problem is partially observed because at any given time instance what another radio tells you (from your radio's perspective) is most likely not the truth, i.e. they are not using exactly that frequency or bandwidth. Hence, any one agent doesn't know the full state of the system and necessitates a POMDP model of some kind. These issues are currently being investigated by researchers in our lab.

The first major result for single-agent POMDPs is in \cite{smallwoo1973optimal}, for a finite horizon problem with a finite state MDP. A paper by Denardo \cite{denardo1967contraction} lays down some important mathematics to prepare for the extension to the infinite horizon. Particularly, he develops contraction and monotinicity properties as well as a convergence criterion based on these properties that would be applied to the cost-to-go functions. The properties developed in \cite{smallwoo1973optimal} do not carry over to the infinite horizon paper on POMDPs \cite{sondik1978optimal}. These classical papers are reviewed more thoroughly in the next section.

\label{Properties}
A brief review of POMDP terminology and properties follow. Assume the internal dynamics of the system are modeled by an $n$-state MDP. At each stage $k$, the goal is to pick a control $u\in \mathcal{U}$ to optimize the performance of the system over its remaining life. To solve this control problem, we encode the current state information in a vector, called the belief state:



\begin{definition}
The belief state at stage $k$ is a vector of state probabilities, $\mathbf{\pi}_k = [\pi_k(1),\ldots,\pi_k(n)],$
where $\mathbf{\pi}_k(i)$ is the probability the current state is $i$ at stage $k$.
\end{definition}
This encoding transforms the discrete-state MDP into a continuous-state MDP. After using control $u$ and observing output $y$, the belief state is updated according to the formula below. The belief state update for some state $j\in \mathcal{X}$:

$$\pi_{k+1}(j|u,y) = \frac{\sum\limits_{i}\pi_k(i)p_{ij}(u)q_{jy}(u)}{\sum\limits_{i}\sum\limits_{l}\pi_k(i)p_{il}(u)q_{ly}(u)}.$$

Now, let's introduce notation for this updating of information or belief-state operation as $\mathbf{\pi}_{k+1}=T(\mathbf{\pi}_k|u,y)$. Next, I present a DP solution to this problem; Define $J_k(\mathbf{\pi}_k)$ as the reward-to-go function, which is the maximum expected reward the system can accrue during the remaining stages starting from stage k, starting from belief state $\pi_k$. Let $g(i,u,y)$ be the generic state, input, and output instantaneous reward, which does not vary across stages. 



Then,
$$ J_k(\mathbf{\pi}_k) = \max\limits_{u\in U} \sum_{i=1}^n \pi_k(i)\sum_{j=1}^n p_{ij}(u)\sum_{y=1}^m q_{jy}(u)[g(i,u,y) + J_{k+1}[T(\mathbf{\pi}_k|u,y)]] .$$ 
The expression above can be further simplified above as:

$$ J_k(\mathbf{\pi}_k) = \max\limits_{u\in U}\left[\sum_{i}\pi_k(i)r^k_i(u) + \sum_{i,j,y}\pi_k(i)p_{ij}(u)q_{jy}(u)J_{k+1}(T(\mathbf{\pi}_k|u,y))\right],~ \forall k \in \{0,1,\ldots,N-1\},$$

where the expected immediate reward for state $i$, at stage $k$ if alternative $u$ is used by $$ r^k_i(u) = \sum_{j}\sum_{y}p_{ij}(u)q_{jy}(u)g(i,u,y). $$


The terminal stage is when $k=N$. If $r_i^N$ is the expected value of terminating the process in state $i$, then $J_N(\pi_N) = \sum_{i}\pi_N(i)r_i^N = \left<\pi_N, r^N\right>$. Note also the reward-to-go function can be rewritten in matrix vector notation for the arbitrary $k$ stage. Define $Pr(y|\pi,u)$ as the probability of observing output $y$ if the current belief-state is $\pi$, and the control $u$ is selected. Then, the recursion becomes: $$J_k(\pi_k) = \max\limits_{u\in U}\left[\left<\pi_k,r^k(u)\right> + \sum_{y\in\mathcal{Y}} Pr(y|\pi,u)J_{k+1}(T(\pi_k|u,y))\right], $$ valid for $k\in\{0,1,\ldots,N-1\}$. This is a DP equation with a continuous state-space, i.e., the $\pi$'s, a finite control space, and a finite observation space. 

Although the DP equation is quite complex looking, it turns out the solution $J_k(\pi_k)$ is piece-wise linear and convex, \cite{smallwoo1973optimal}: 

\begin{theorem}
Piece-wise Linear and Convex Cost-to-go Function:

$$J_k(\pi_k) = \max\limits_{l}\left[\sum_{i=1}^n \alpha^{(l)}_k(i)\pi_k(i)\right],$$ for some set of alpha vectors $\alpha^{(l)}_k = [\alpha^{(l)}_k(1), \alpha^{(l)}_k(2),\ldots,\alpha^{(l)}_k(n)]$, for $l=1,2,\ldots$ at stage $k$.
\end{theorem}

The proof is done by induction. Note that at time $k=N$, $$J_N(\pi_N) = \max\limits_{u\in U}\left<\pi_N,r^N(u)\right>,$$ has the desired form of being piece-wise linear in $\pi$. The proof is important because it allows one to recursively compute the optimal reward function and hence the optimal policy. See \cite{smallwoo1973optimal} for the completion of the proof. 





In the beginning, $A_N = \{r^N(u): u\in \mathcal{U}\}$. For the earlier stages, the alpha vector set can be constructed as follows: 
$$ A_k=\left\{\frac{r_k(u)^T}{m} + P(u)Q_y(u)\alpha^T:~\alpha\in A_{k+1},~u\in U,~y\in Y\right\},~ \forall k \in \{0,1,\ldots,N-1\}. $$
Once $A_k$ is constructed from stage $k$ computations, the optimal policy $\mu^*_k(\pi_k)$ can be computed as follows: $$ \mu^*_k(\pi_k) = u(\arg\max\limits_{\alpha\in A_k} < \alpha, \pi_k>).$$


All exact methods stem from the some of the classical results studied in this section and the classical papers studied in the next section, particularly the work of Sondik. At each stage $k$ one iteratively constructs a set of alpha vectors $A_k$ from the future stage. In \cite{krishnamurthy2016partially},
a general incremental pruning algorithm is proposed to reduce complexity while maintaining optimality by eliminating $\alpha$ vectors that never contribute to the cost-to-go function. The paper by Monahan, \cite{monahan1982state}, computes the set of $\alpha$ vectors $A_k(u)$ at each stage which has $|U||A_{k+1}|^m$ alpha vectors in it, where $m$ is the number of possible observations, and the other parts are cardinalities of finite sets. These are pruned according to the linear program in \cite{krishnamurthy2016partially}. Monahan also goes over an enumeration strategy for value iteration for POMDPs, but in general, as we have already seen, the problem grows exponentially in the number of alpha vectors used to approximate the value function. This necessitates the use of pruning strategies to reduce complexity and hence approximate computation for problems with large state, action or observation spaces. Approximation methods for the POMDP are presented after a review of classical papers in the next section. Hence, if one only wants to read or learn about the main approximation methods for POMDPs, the next section can be safely skipped. 

The reading road-map of this section is start with \cite{bertsekasdynamic1} Ch.4 Problems with Imperfect State Information. There, he illustrates well the ever increasing computational complexity associated with an increasing horizon length for POMDPs, but this chapter is not great for exploring the introductory material of POMDPs. Skip 4.1-4.2., and read 4.3 Sufficient statistics, paying particular attention to 4.3.2 Finite-State Systems. This chapter begins to illustrate where assumptions of MDPs and dynamic programs breakdown when perfect state information is not available for decision making. After reading this chapter, an appreciation for POMDPs can be developed further by reading \cite{krishnamurthy2016partially}. The textbook \cite{krishnamurthy2016partially} serves as a better introduction to POMDPs and for entering recent research directions in the later chapters, e.g., social sensing. After this section is completed, I move to focus entirely on approximation methods for the POMDP. 


\section{Classical Papers for POMDPs}
\label{classical}

\subsection{Paper 1: Astrom, ``Optimal Control of Markov Processes with Incomplete State Information"}
\subsubsection{Overview of the paper}
I start my survey of classical papers with Astrom's paper because it closely aligns with the goal of making optimal decisions on problems with countable state spaces when state information is incompletely known. Astrom solves the problem of optimal control with incomplete state information using the mature field of DP. The state spaces and control spaces in this paper are more general than what has been considered so far in this survey paper. As well as presenting unified notation, I convert theorems when relevant into our finite-space setting, possibly infinitely countable, but never uncountable. A preliminary paper that may be read before reading this paper is  \cite{stratonovich1965conditional} because a result from section III of it is used, equation 3.19. Another paper, that may be of interest is \cite{dynkin1965controlled} as it deals with control of control of a partially observed random sequence; however, it is more relevant as a setting to bandit type problems. It is not required reading for this POMDP survey, but suggested. At the end of each section in the rest of the survey, I provide a few paragraphs with pointers to further literature, related to the papers considered in that section.


Summary: the beauty of \cite{aastrom1965optimal} is to introduce control to Markov processes, and find an optimal controller that uses all of the observations available up to the current time or stage. Instead of postulating a form of a control law as is done in some areas of control theory, he minimizes a specific cost functional. The design of his problem as a stochastic variational problem (svp) leads to a classical two step procedure as is known for solving optimal control of deterministic linear systems with a quadratic cost. These steps are 1) estimate the state based on observations, and 2) calculate the control variable from the estimated state. This paper presented five novel theorems at the time of publication. I summarize the results and offer new ideas where holes are found. 


The cost-to-go or performance measure of the system is a functional of the state trajectories and controls selected. We seek to minimize this functional in expectation. It is noted that the selected controls may depend on the history of the measurements up to the current time $k$. To solve his more general problem of dynamics resulting from a stochastic differential equation, he discretizes the state and time variables to be able to use Markov chain theory. The theory for the quantized problem and its solutions are presented, and it is solved using DP. The last two theorems form provable lower, and upper-bounds for the solution. This leads to a notion of the value of complete state information and partial or incomplete state information relative to no state information. This is interesting, because this leads to a connection of information theory with stochastic optimal control.

\subsubsection{Set up of the POMDP Theory and Statement of the Problems}

Denote the countable space Markov chain by $\{X_k,~k=0,1,\ldots\}$. For simplicity assume a finite state space $\mathcal{X} = \{1,2,\ldots,n\}$ and discrete time $\mathbb{Z}\cup 0$. We have an initial prior pmf on the initial state $b_0(i) = Pr(X_0=i)$. The possibly time-varying controlled state transitions is denoted by $P(u,k)$ with the same constraints on $P$ given earlier in the survey. Here, however $u$ is a member of a compact set $U$, referred to as the set of admissible controls. Keep in mind, most of the time for computer-based systems $U$ is finite. In his model, the transitions are continuous functions of $u(k)$.

Also, the \say{output} of the system or the observable process is denoted by $\{Y_k,~k=1,2,\ldots\}$, which is a discrete time Random Process. Most of the time the observation space is finite: $\{1,2,\ldots,Y\}$. The observation transitions go in a matrix $Q$, and each element $q_{ij} = Pr(Y_k=j|X_k=i)$ with the following conditions met: $\ q_{ij} \geq 0,~\sum\limits_{j} q_{ij} = 1$. The matrix $Q$ reflects the probabilities of measurement errors. Perfect state information results when $Q=I$, the identity matrix, and $Q$ may not be square in general. Denote as a realization of this process, i.e. the measurements up to time $k$ by $\underbar{Y}(k) = [y_1, y_2,\ldots,y_k]^T$. Define a set of control functions as $C = \{\mu(y_1,y_2,\ldots,y_k, k): k=1,\ldots,N\}$, and refer to it as a strategy or control law. The control law is admissible iff $\mu\in U\ \forall k$. This introduces feedback into the problem as it directly relates the measurements or observations with the control to be applied. Define the instantaneous cost function, $g(x,u,k)$, to be a scalar function of $x,u,k$ that depends continuously on $u$. The total cost in expectation is $$ E\{L\} = E\biggl\{\sum_{k=1}^{N}g(X_k,\mu(y_1,\ldots,y_k),k)\biggr\}.$$
Problem 1 in \cite{aastrom1965optimal}:
\begin{displayquote}
Find an admissible control signal whose value at time $t$ is a function of the outputs observed up to that time and are such that the expected value of the total cost is minimal.
\end{displayquote}
Solution to Problem 1:
We seek to minimize this expectation over all possible choices of $u$:
$$ \min\limits_{u_1}\ldots\min\limits_{u_N} E\biggl\{\sum_{k=1}^{N}g(X_k,u_k,k)\biggr\}.$$ $u_k$ is a function of $(y_1,\ldots,y_k)$, the observations up to time $k$. The solution is given by DP. At time $k=N$,  $\underbar{Y}(N) $ is obtained and we must determine $u_N$ as a function of the observations. At the final stage, we must minimize $ \min\limits_{u_N} E\{g(X_N,u_N,N)\}$. We can rewrite this expression using the smoothing property of conditional expectation to make the dependence on $\underbar{Y}(N)$ explicit: $$ E_{\underbar{Y}(N)}\{\min\limits_{u} E\{g(X_n, u, N)|\underbar{Y}(N)\}\} .$$ Define the optimal cost-to-go function by: $$ J_N = \min\limits_{u} E\{g(X_n, u, N)|\underbar{Y}(N)\},$$ and note the dependence of $J_N$ on all of the observations, which comes from the conditional distribution of $X_N$ given $\underbar{Y}(N)$. To emphasize this point, define $$\pi_N(i) = Pr(X_N=i|\underbar{Y}(N)), \textit{ and } \pi_N = [\pi_N(1),\ldots,\pi_N(n)]^T$$ for finite states, or

$\pi_N = [\pi_N(1),\pi_N(2),\ldots]^T$ for a countably infinite state space. Thus, the minimal cost at stage $N$ is $E_{\underbar{$Y$}(N)}\{J_N(\pi_N)\}$, where $J_N= J_N(\pi_N)$.

It is shown recursively in \cite{aastrom1965optimal}, that the minimal cost of the remaining $N-k$ stages can be written as:
$$ \min\limits_{u_{k+1}} \ldots \min\limits_{u_N} E\biggl\{\sum_{t=k+1}^{N}g(X_t,u_t,t)\biggr\} = E_{\underbar{Y}(k+1)}\{J_{k+1}(\pi_{k+1})\}. $$ Using this hypothesis, we have $$ \min\limits_{u_{k}} \ldots \min\limits_{u_N} E\biggl\{\sum_{t=k}^{N}g(X_t,u_t,t)\biggr\} = \min\limits_{u_{k}} E\biggl\{g(X_k, u_k, k) + E_{\underbar{Y}(k+1)}\{J_{k+1}(\pi_{k+1})\}\biggr\}.$$ Hence, the optimal control has to be a function of $y_1,\ldots,y_k$ to minimize the above expression. One more step makes this explicit using the smoothing property: $$ = E_{\underbar{Y}(k)}[E\{g(X_k, u_k, k)|\underbar{Y}(k)\} + E\{J_{k+1}(\pi_{k+1})|\underbar{Y}(k)\}], $$ which can be expressed further using the two conditional densities of the state $X_k$ given $\underbar{Y}(k)$ and the conditional density of $Y_{k+1}$ given $\underbar{Y}(k)$.

An application I thought of for this was taking this result and applying it for a finite state, and finite observation system. The dynamic program at stage $k$ would minimize over $u_k$ the following expression using now two conditional pmfs:
$$ E_{\underbar{Y}(k)}[\sum_{x\in\mathcal{X}}g(x,u_k,k)P(x|\underbar{Y}(k)) + \sum_{y\in\mathcal{Y}}J_{k+1}(\pi_{k+1})P(y|\underbar{Y}(k))].$$ Some interesting Markov assumptions could be made about these two conditional pmfs that would make this a much more useful formula for online computation than the original continuous integrals involving densities. However, the continuous integrals may be useful for developing some soft-decision decoders for error control coding research.

Next, he derives a recursive filter for the state estimator $\pi_k$ that updates after observation $y_k$ is available; this is our optimal Bayesian filter from before. Refer to the derivation in the paper for more details. Using $P, Q,$ and  $\pi_k$ definitions we can build the following recursive update: $$ \pi_{k+1}(i) = \frac{\sum\limits_{s\in\mathcal{X}} \pi_k(s)p_{si}(u)q_{ij} } {\sum\limits_{s\in \mathcal{X}} \sum\limits_{\ell \in \mathcal{X}} \pi_k(s)p_{s\ell}(u)q_{\ell j}},$$ where $y_{k+1} = j$.

Introduce the notation $z_{ij}(u,\pi_k) = \sum\limits_{s\in\mathcal{X}} \pi_k(s)p_{si}(u)q_{ij}$. Notice that $z_{ij}(u,\pi_k) \geq 0$ and $j$ refers to the observation. Introduce the vector $z^j = [z_{1j}, z_{2j},\ldots, z_{nj}]$ and define the norm on this vector as $||z^j|| = \sum\limits_{i\in\mathcal{X}} |z_{ij}|$. The norm is interpreted as the conditional probability $$||z^j|| = P(Y_{k+1}=j|Y_1=y_1,\ldots,Y_k=y_k).$$
Now, we can revisit the DP equation and simplify its expression:
$$ E\biggl[g(X_k,u_k,k) + E_{\underbar{Y}(k+1)}\{J_{k+1}(\pi_{k+1})\}\biggr] = E_{\underbar{Y}(k)}\biggl\{\sum_{x\in\mathcal{X}} g(x,u_k,k)\pi_k(x) + \sum_{j\in\mathcal{Y}} J_{k+1}(\frac{z^j(u,\pi_k)}{||z^j(u,\pi_k)||})||z^j(u,\pi_k)||\biggr\} .$$

Hence, $$ \min\limits_{u_{k}}\ldots \min\limits_{u_N} E\biggl\{\sum_{t=k}^{N}g(X_t,u_t,t)\biggr\} = E_{\underbar{Y}(k)}[J_k(\pi_k)], $$ where, $$ J_k(\pi_k) = \min\limits_{u_k}\biggl\{\sum_{x\in \mathcal{X}} g(x,u_k,k)\pi_k(x) + \sum_{j\in \mathcal{Y}} J_{k+1}(\frac{z^j(u_k,\pi_k)}{||z^j(u_k,\pi_k)||})||z^j(u_k,\pi_k)||\biggr\}.$$ The induction follows trivially to any arbitrary stage. Now that the intuition is set, and the basic building blocks are assembled, I present the theorems in order.

\subsubsection{The Theorems}
Theorem 1: Let the control law 
$ \underbar{$\mu$}^o = \{\mu^o(\pi_k, k), k=1,\ldots,N\} $ minimize the functional 
$ E\{L\} = E\{\sum\limits_{k=1}^N g(X_k, \mu(y_1,\ldots,y_k), k)\}. $

Then, for any stage k: $$J_k(\pi_k) = \min\limits_{u_k} E\biggl[\ldots\min\limits_{u_N}E\biggl[\sum_{t=k}^N g(X_k, u_k, k)|~\underbar{$Y$}(N)\biggr]\ldots|\  \underbar{$Y$}(k)\biggr] = \sum_{t=k}^N\sum_{i}g(i,\mu^o(\pi_t,t),t)\pi_t(i).$$

Theorem 2: Let the functional equation for $J_k(\pi_k)$ have a solution $J_t(\pi_t)$. Then, the Problem 1 has a solution, and the control law $\mu^o$ minimizes the functional $E\{L\}$. Its minimal value is $E_{y_1}[J_1(\pi_1)]$. Another interesting fact from the proof of this theorem is that the continuity of $g(x,u,t), p_{ij}(u)$ implies that if there is a solution to the cost functional, say $J$, then this solution is continuous in $\pi$.

Some more discussion of the above two theorems yields more insight. The cost functional that is being minimized, can be solved apriori knowing only the problem data $g, P, \textit{ and } Q$ without seeing any observations $\{y_1,y_2,\ldots\}$. The controller took the form $u=u(\underbar{Y}(k),k) = u(\pi_k(\underbar{Y}(k)),k)$. Finally, note that the dependence of $u$ on $\{y_1,y_2,\ldots\}$ only enters from the conditional distributions $\pi_k$. The function $u$ can be obtained from minimizing the cost functional, and can be obtained offline. The function $\pi_k=\pi_k(\underbar{Y}(k))$ expresses the conditional distribution of the state as functions of the observed data $\underbar{Y}(k)$, and is given by a recursion already presented. 


The one lemma given in \cite{aastrom1965optimal} is also quite useful. He proves that for a control law such that $u_k = \mu(\pi_k,k)$, the set of conditional probabilities $\{\pi_k: k=0,1,2,\ldots\}$ is a Markov chain. Proof: the optimal control law $u_k$ is a function of $\pi_k$ and the equation 

$$ \frac{z^i(u,\pi_k)}{||z^i(u,\pi_k)||} \implies P(\pi_{k+1}|\pi_k,\pi_{k-1},\ldots,\pi_1) = P(\pi_{k+1}|\pi_k). \implies \pi_k \textit{ is a M.C. } \blacksquare.$$

Next, theorem 3 is presented which is based on an adjusted statement of problem 1. He considers a variational problem relative to the $\pi$-process. Let $g(u,k)$ be the vector $g(u,k) = [g(1,u,k),\ldots,g(n,u,k)]^T$. Introduce the new functional as $ E\{\sum_{k=1}^N\langle g(u,k),\pi_k\rangle\}$ where $\langle a,b \rangle$ denotes an inner product between two vectors $a,b$. $E$ denotes expectation with respect to $\{\pi_1,\ldots,\pi_N\}$. Now, consider the Problem 2 statement:
\begin{displayquote}
Find a sequence of admissible controls $\{u_k: k=1,\ldots,N\}$ s.t. our functional is minimal. The value of $u$ at any time $k$ may depend on all previous and current $\{\pi_1,\ldots,\pi_k\}$.
\end{displayquote}
This gives rise to Theorem 3: 
\begin{displayquote}
The problems 1 and 2 are equivalent in the sense that if one of the problems has a solution, then the other problem also has a solution. Furthermore, the optimal control law is $u_k = \mu^o(\pi_k,k)$ in both cases where $\mu^o$ is given by Theorem 1. In problem 2, there is complete state information, and so it reduces to a kind of classical DP recursion. Lastly, Theorem 3 implies that the problem of optimal control of a Markov chain with incomplete state information can be transformed into a problem of optimal control of a process with complete state information. However, the \say{state space} of the problem with complete information is the space of probability distributions over the states of the original problem.
\end{displayquote}

I believe this is the first instance in the research literature of what is today known as \say{belief space}, and note the observation space is finite, with at most m points. However, in this formulation, the control is not assumed to affect the observation. The division of controlling a Markov chain with incomplete state information can be broken into two parts.
Part1: solve either Problem 1 or Problem 2. This gives the optimal controller $u=u(\pi,k)$. Part2: calculate the conditional probability distribution $\pi_k$ of the states of the associated M.C. from the measured signal $\underbar{Y}(k)$.

The last two theorems of the paper provide lower and upper bounds on the optimal cost-to-go functions. Two cases are presented. The complete state information case provides a lower bound on the optimal cost-to-go function, and the open-loop system using only apriori information provides an upper bound on the optimal cost-to-go function. Case1(Complete state info): Here, $Q=I$, and define the cost-to-go with complete state information as $S_k$ from stage $k$. Then, $$S_k(i) = \min\limits_{u}\biggl[g(i,u,k)+\sum_{j}p_{ij}(u)S_{k+1}(j)\biggr].$$ Case2(Open-Loop system, i.e., no state information): the other extreme is when no aposteriori state information is available. Assume $q_{ij}=c \textit{ constant }~\forall i,j$ . Then, we get $z_{ij} = c\sum\limits_{s}\pi_k(s)p_{si}(u)$, and $\sum\limits_{i}z_{ij} = c$. Also, in this case $\pi_{k+1}(i) = \sum\limits_{s}\pi_k(s)p_{si}(u)$. The conditional probabilities are independent of $\underbar{Y}_{k+1}$, which means the measurements contain no information for calculation of $\pi_{k+1}$. Rewriting the cost functional for this case gives us:
$$ V_k(\pi_k) = \min\limits_{u}\biggl\{\sum_{i}g(i,u,k)\pi_k(i) + V_{k+1}(P^T(u)\pi_k)\biggr\},$$ which appears to be using the C-K equation presented earlier in the survey to run the open loop minimization.

Theorem 4: 
\begin{displayquote}
Let the solution of the imperfect-state cost problem be: $$J_k(\pi_k) = \sum_{t=k}^N\sum_{i}g(i,\mu^0(\pi_t,t),t)\pi_t(i) \text{ be given by } J_k(\pi_k) ,$$ and that of case 1) by $\underbar{J}_k(\pi_k)$. Then, it can be shown that $\underbar{J}_k(\pi_k) \leq J_k(\pi_k)$. 
\end{displayquote}

Theorem 5: 
\begin{displayquote}
Let the solution of the functional equation in Theorem 4 be $J_k(\pi_k)$, and that of case 2) open-loop system be $V_k(\pi_k)$. Then, it can be shown that $J_k(\pi_k) \leq V_k(\pi_k)$. 
\end{displayquote}

Some closing remarks and ideas: from the above two theorems one has a final beautiful insight. $E_{y_1}[V_1(\pi_1) - \underbar{J}_k(\pi_k) $ is taken as the value of having perfect state info relative to none at all, and $E_{y_1}[J_1(\pi_1) - \underbar{J}_k(\pi_k) ]$ is the value of having incomplete state information relative to none at all. It would be interesting to follow this point up, and investigate other upper and lower bounds than the ones presented. Is it possible to prove tighter upper and lower bounds? Are there better approximations for these bounds? In particular, for the lower bound, if one was found with domain the space of beliefs that would be a theoretical advance. This lower bound is a linear function of $\pi_k$. So, if we found a lower bound to $J_k$ in the space of beliefs, that would mean we have found a better algorithm than the current DP solution at minimizing the total costs over the finite horizon. A tighter upper bound in the belief space would also be interesting, because that would mean we are really close to being optimal, and this may allow us to save on a lot of unnecessary computation to reach that bound. Additionally, as a major research hole, the number of times observations or \say{measurements} are made could be incorporated into the instantaneous cost. Then, one could assign a cost to running a measurement, and determine when is it optimal to make a measurement. I would call this an optimal measuring policy. Finally, is it possible to define a notion of entropy on the transition matrix $P(u)$ or with the belief-state sequence? This could lead to a new type of maximum exploration strategy, and possibly help with creating new upper and lower bounds for the optimal-cost-to go functions. The research area of exploration appears completely unexplored for POMDPs. Information theory may prove as a most valuable tool for working with the belief states as the distributions of interest and using some of Shannon's information measures, whether relative entropy or entropy to investigate exploration strategies for POMDPs.

\subsection{Paper 2: Smallwood and Sondik, ``The Optimal Control of Partially Observable Markov Processes Over a Finite Horizon"}

\subsubsection{Overview of the paper}
This paper \cite{smallwoo1973optimal}'s main results were reviewed in \autoref{pomdp}, but here I provide more of an overview of the paper and do not repeat what has been said previously about the paper. This paper formulates the optimal control problem specifically for a finite-state discrete-time Markov Chain (MC). The motivating example of this paper that illustrates the theory developed is a machine maintenance example that is over-viewed in this section of the survey, and helps put the theorems, proof of theorems, and other mathematical results in their context. Typically, in DP or operations research, cost minimization is pursued, but here for this paper, reward maximization is pursued, as in RL.  $J_k$ is the optimal reward-to-go function from stage k, whereas typically this is called the optimal cost-to-go function from some state.

Moreover, the machine example serves also as an analogy for POMDPs. This is true because while a machine operates one cannot generally know the internal states exactly by only inspecting the outputs. For this example all one can observe is the quality of the finished product. Only a finite number of observations are considered. The machine produces exactly one finished product once an hour, at the end of the hour. The machine is run by two internal components that may fail independently at each stage $k$. If a component has failed, it increases the chance that a defective product will be produced.

Assume that the finished product is either defective or not defective. If the finished product is defective, the system incurs a reward $r=0$. While if the finished product is not defective, it incurs a reward of $r=1$. For this problem there are three states that could be labeled as $x_1=0, x_2=1, x_3=2$. These states correspond to zero, one, and both parts inside of the machine being broken inside the machine, respectively. The two observations are $y_1= D, y_2=\bar{D}$ which correspond to seeing a defective or not defective product at the end of any given hour on the factory floor. Refer to Figure 1 in \cite{smallwoo1973optimal}, which illustrates the relation between the internal Markov process with $n=3$ states and the $m=2$ possible observations via the transition diagram.

Furthermore, Sondik uses different terminology for the controls language we have been using. He calls a control or decision a \say{control alternative}, as in another option to choose from a finite set of alternatives. For this problem there are four such control alternatives which are the distinct ways to run the manufacturing process from the start of each hour. These alternatives are just admissible controls in $U$ in our framework. 

The control options are:
\begin{enumerate}
  \item  Continue operation, i.e., do not examine finished product,
  \item  Examine product at the end of hour,
  \item  Stop machine for the hour, disassemble it, inspect the two internal components, and replace any that have failed,
  \item  The last option is to use the whole hour to just replace both components with no inspection.
\end{enumerate}

He discusses the costs for this problem at the end of the paper, but it is not hard to see each control will cost a different amount at each stage. Hence, we must determine an optimal policy that maximizes the total reward. Another notation he uses in the paper for an instantaneous reward term, which is most generic: $W_{ij\theta}^a$, includes the possibility of rewarding based on observations too. In our framework and notation we could write: $c(i,u,y,k)$ for an observation dependent instantaneous cost, that also incorporates stage number. The goal is to find the optimal policy among these 4 options, each having different rewards associated with them, and knowing the complete history of the machine's operation.

In table I of \cite{smallwoo1973optimal}, other POMDP application examples of interest are presented along with their modeling assumptions. I only mention two examples here related to controls and communications. Here, I present a few examples with the states, observations, and controls. For the first example it shows the need for considering more control options, and even using modern day encoding and decoding techniques, while the second considers a target tracking application.

\begin{enumerate}
   \item Decoding of Markov Sources
   \begin{itemize}
     \item State: status of the source,
     \item Observation: output of the noisy channel,
     \item Control Alternative: not considered.
   \end{itemize}

   \item Search for a moving object
   \begin{itemize}
     \item State: status of target object,
     \item Observation: result of search,
     \item Control Alternative: level of expenditure of search resources.
   \end{itemize}

\end{enumerate}


Another application area mentioned in \cite{smallwoo1973optimal} is that of target tracking. I have a few ideas of furthering the ideas from the papers \cite{pollock1970simple},\cite{eagle1984optimal} that studied tracking a Markovian moving target. The seminal paper on the optimal detection research question can be read in \cite{pollock1970simple}. The main working questions in \cite{pollock1970simple} are 1) What is the minimum expected number of looks required for detection of the Markovian moving target and 2) given a finite number n of looks, what is the maximum probability of detection that can be achieved. In \cite{pollock1970simple}, only a two state POMDP model was developed, and the goal was to maximize the probability of detection over a finite number of stages using DP methodology. The paper \cite{eagle1984optimal} extended that approach to include an arbitrary finite number of $n$ states for the same goal of maximizing the probability of detection. The paper \cite{eagle1984optimal} addresses the call to action from the seminal paper by explicitly formulating the search problem as a POMDP. This paper makes use of the POMDP solution techniques over a finite horizon by utilizing ideas contained in\cite{smallwoo1973optimal}, and \cite{monahan1982state}. Lastly, for the literature review, \cite{singh2003optimal} extended \cite{eagle1984optimal}'s research on the optimal detection problem to the infinite-horizon case.

A Markovian moving target can be considered with momentum for future research directions. The system could have state vectors that are mixtures of completely observable and partially observed states. Effectively, I would like to develop a POMDP model for tracking this moving target with communication models of the received state information. In \cite{pollock1970simple},  a two-state moving target model was formulated, which can be converted to a 3 state process, where the third state signifies detected or not detected.

\subsection{Paper 3: Denardo, ``Contraction Mappings in the Theory Underlying Dynamic Programming"}

\subsubsection{Overview of the paper}
This paper ties together a variety of disparate DP problems under a general framework by noticing the many similarities between all the different models in this era. The math developed in this paper helps in studying any infinite horizon problem: POMDP or not. The main properties most of these models possess are:
\begin{enumerate}
  \item Contraction,
  \item Monotonicity,
  \item N-stage contraction.
\end{enumerate}

The property 3. is actually a weakened statement of the first property. These properties surfaced in the works of famous authors such as \cite{howard1971dynamic}, \cite{blackwell1965discounted}, and \cite{bellman1957dynamic}. In the paper \cite{denardo1967contraction}, the term \say{policies} is introduced; denote it as $\mu$, and call the return or the reward-to-go function $J_{\mu}$. Next, he introduces a maximization operator, denoted as $A$, and shows it has the contraction property. He notes that, \say{The fixed-point theorem for contraction mappings assures us that the equation $AJ=J$ has a unique solution $J^*$.}. Define the optimal return or optimal reward-to-go function as $J_{opt}$. $J_{opt} = \sup\limits_{\mu} J_{\mu}$. If both contraction and monotonicity assumptions are satisfied it is the case that $J^* = J_{opt}$, i.e., the fixed point is the optimal return function. A sufficient condition is given for a policy to obtain $J^*$.

The author provides similar results for assumptions 2-3. $\epsilon$-optimal policies are discussed which are near optimal policies, but not optimal. Some computational techniques for approximating $J^*$ are given, namely using the idea of successive approximation to build a Value Iteration Algorithm (VIA), and extending the work in \cite{howard1971dynamic} on Policy Iteration Algorithm (PIA). The properties already mentioned are exploited in these algorithms. Section 8 gives many varieties of models and mixtures of different modeling assumptions. For this survey, example $1$ matters the most for the infinite horizon in preparing for Sondik's paper on the topic of POMDPs for the infinite horizon. At the very end of this paper, nice definitions of metric spaces, contraction mappings, and the fixed-point theorem are given.

\subsubsection{Contraction Mappings}

Introduce some general notation first. Let $\Omega$ be a set. An element of $\Omega$, denoted by $x$ is a point of this set. The set of decisions is denote by $U(x)$. An element of $U(x)$, denoted as $u_x$ is called a decision. The policy space $M$ is given by the Cartesian product of the decision sets: $M = \times_{x\in \Omega}~U(x)$. An element of $M$ is referred to as a policy, and denoted by $\mu$ in most of the literature. It is a mapping from state-space to action-space. Next, to introduce the return function or the return-to-go function as I like to call it. Denote the collection of all bounded functions from $\Omega \to \mathbb{R}$ by $V$, i.e.,
 $$J\in V \textit{ iff } J: \Omega \to \mathbb{R},~\sup\limits_{x\in \Omega}|J(x)| < \infty.$$ Next, define a metric on V between two return-to-go functions as $\rho(J,J') = \sup\limits_{x\in \Omega}|J(x) - J'(x)|.$ It is noted that the metric space $V$ is complete.

 Define the \say{return} $h$ as a function. Given a triplet $(x, u_x, J)$ with $x\in \Omega, u_x\in U(x), J\in V$. The function $h(x,u_x,J)$ is thought of as the total payoff starting at $x$, choosing action $u_x$, with possibly probabilistic transition to some point $z\in\Omega$, and thus receiving $J(z)$. Thus, the pair $(x,u_x)$ causes such a transition to a next point $z$. Here points would be states for our applications, but this paper is the most mathematical surveyed so far, most likely in all of the survey. $h(x,u_x,\cdot)$ describes what $(x,u_x)$ yields as a function of $J$. Now, define the contraction assumption:
 \begin{displayquote}
 For some $0\leq c < 1$, $|h(x,u_x,J)-h(x,u_x,J')| \leq c\rho(J,J')$ for each $J\in V,\  J'\in V,\  x\in\Omega,\  u_x\in U(x)$.
 \end{displayquote}
 
Let the policy be denoted as $\mu(x)$ where the decision $u_x$ is applied when at the point $x$. For each $\mu\in M$, a function $H_{\mu}$ having domain $V$ and range contained in $V$, i.e. $H_{\mu}: V \to V$ is defined by: $[H_{\mu}(J)](x) = h(x,u_x,J),$ where $H_{\mu}(J)\in V$ which $H_{\mu}$ assigns $J$ to this next bounded function. $[H_{\mu}(J)](x) \in \mathbb{R}$ which the function $H_{\mu}(J)$ associates with the point $x$. The contraction assumption is equivalent to the following.
 For some $0 \leq c < 1$, $\rho(H_{\mu}J, H_{\mu}J') \leq c\rho(J,J')$ for each $J, J'\in V$ and $\mu \in M$. Hence, it is noted that \say{$H_{\mu}$ is a contraction mapping, and the fixed-point theorem for contraction mappings guarantees that $H_{\mu}$ has a unique fixed point, i.e., $J_{\mu}$.}. For each $\mu\in M$ a unique element $J_{\mu}\in V$ exists such that $J_{\mu}(x) = h(x,\mu(x),J_{\mu}) \forall x\in\Omega$. $J_{\mu}$ is called the \say{return function}; out of reverence for Dimitri Bertsekas, I call it the reward-to-go function as before in the finite horizon setting, in connection with the cost-to-go function. The optimal reward-to-go function $J_{opt}(x) = \sup\limits_{\mu\in M}J_{\mu}(x)$.

 Theorem 1: 
 \begin{displayquote}
 Suppose the contraction assumption is satisfied. For any $\mu\in M$, and any $J\in V$ we have $\rho(J_{\mu},J) \leq \frac{\rho(H_{\mu}J, J)}{(1-c)} $.
 \end{displayquote}
 
By usage of the Triangle inequality we have:
\begin{equation*}
\begin{split}
\rho(H^n_{\mu}J, J) &\leq \sum_{i=1}^n \rho(H^i_{\mu}J, H^{i-1}_{\mu} J)\\
                    &\leq \sum_{i=1}^n c^{i-1}\rho(H_{\mu}J, J)\\
                    &\leq \frac{\rho(H_{\mu}, J)}{(1-c)}
\end{split}
\end{equation*}

Since,
$$ \lim\limits_{n\to 0} \rho(H^n_{\mu}J, J_{\mu}) \to 0 \implies \rho(J_{\mu}, J) \leq \frac{\rho(H_{\mu}J,J)}{(1-c)}$$



 For a Maximization Operator:
 define a map $A$ having domain $V$ by $(AJ)(x) = \sup\limits_{u_x\in U(x)} h(x, u_x, J)$, for each $J\in V,\  x\in \Omega$. Assume that the range of $A$ is contained in $V$. The next theorem shows that $A$ is indeed a contraction mapping.

 Theorem 2:
 \begin{displayquote}
 suppose the contraction assumption is satisfied. For each $J\in V, J'\in V,$ we then have $\rho(AJ, AJ')\leq c\rho(J,J')$.
 \end{displayquote}

 The author states that, \say{the fixed-point theorem guarantees that $A$ has a unique fixed point, i.e., $\exists$ exactly one element $J^*$ of $V$ such that $J^*(x) = \sup\limits_{u_x\in U(x)} h(x,u_x,J^*)\  \forall x \in \Omega$.}. This is a very general notation for the functional equation of DP in terms of $J^*$.

 Two related questions are whether $J^*$ can be approximated by some other reward-to-go function $J_{\mu}$ for some policy $\mu$ and whether $J^*$ is the optimal reward-to-go function, i.e., whether $J^* = J_{opt}$. Let us first address the first question. The existence of $\epsilon$-optimal policies is given by satisfying $\rho(J_{\mu}, J^*) \leq \epsilon \forall \epsilon >0$. Question two is addressed in section 4, where a sufficient condition for $J^* = J_{opt}$ is given. Without this condition to be introduced later, the fixed point $J^*$ may be less than $J_{opt}$, i.e. $J^* < J_{opt}$ and equality is not obtained.

 Corollary 1: 
 \begin{displayquote}
 For $\epsilon > 0, \exists$ a policy $\mu$ such that $\rho(H_{\mu}(J^*), J^*) \leq \epsilon(1-c),$ and any such $\mu$ satisfies $\rho(J_{\mu}, J^*)\leq \epsilon$ and if $\rho(H_{\mu}(J^*), J^*) = 0$, then $J_{\mu}= J^*$.
 \end{displayquote}


 Proof: For $\epsilon >0$, existence of $\mu$ such that $\rho(H_{\mu}(J^*),J^*)\leq \epsilon(1-c)$, which follows directly from the generic functional equation given earlier:
\begin{equation*}
\begin{split}
J^*(x) &= \sup\limits_{u_x\in U(x)} h(x,u_x,J^*) \\
       &\implies 0 \\
       &= \sup\limits_{u_x\in U(x)}|h(x,u_x,J^*)-J^*| \implies \sup\limits_{x\in \Omega}\{\sup\limits_{u_x\in U(x)}|h(x,u_x,J^*)-J^*|\} \\
       &= \rho(H_{\mu}(J^*),J^*)\\
       &= 0 \leq \epsilon(1-c)
\end{split}
\end{equation*}

Then, using Theorem 1, sub in $J^*$ for $J$ which yields:
 \begin{equation*}
\begin{split}
\rho(J_{\mu},J^*) &\leq \frac{\rho(H_{\mu}J^*, J^*)}{(1-c)} \\
                  &\leq \epsilon\  \forall \epsilon \geq 0.
\end{split}
\end{equation*}
Corollary 2: 
\begin{displayquote}
Suppose for each fixed $x$ that $h(x, \cdot, J^*)$ is a continuous function of $u_x$ in a topology for which $U(x)$ is compact. Then, $\exists$ a policy $\mu$ such that, $J_{\mu}= J^*$. 
\end{displayquote}

Proof: with $B$ as an operator on $V$, define the modulus of $B$ as the smallest number $c$ such that $\rho(BJ, BJ') \leq c\rho(J,J')$ for each $J, J'\in V$. With $I$ as an arbitrary non-empty set, suppose $\{B_{\alpha}: \alpha \in I\}$ is a collection of operators on $V$, each of which has modulus $c$ or less. Define the function $E$ having domain $V$ by $(EJ)(x) = \sup\limits_{\alpha\in I}(B_{\alpha}J)(x)$, and suppose that $E$ has range contained in $V$. Then, an argument similar to Theorem 2 establishes the fact that $E$ has modulus $c$ or less.

The assumption below suffices for $J^*, J_{opt}$ to be identical. For $J, J'\in V,$ we write $J \geq J'$ if $J(x)\geq J'(x)\  \forall x \in \Omega$.

The Monotonicity Assumption: If $J > J'$, then $H_{\mu}(J) \geq H_{\mu}(J') \forall \mu\in M$. See Denardo's dissertation \cite{denardo1965sequential} for more on monotinicity. Many reward-to-go functions satisfy the monotonicity assumption. It is equivalent to $h(x,u_x,J)>h(x,u_x,J')$ if $J\geq J'$. 

Theorem 3: 
\begin{displayquote}
Suppose the monotonicity and contraction assumptions are satisfied. Then, $J^*=J_{opt}$.
\end{displayquote}

Proof: From Corollary 1 we know $J^*\leq J_{opt}$. Since $H_{\mu}J^* \leq J^*$ for each $\mu$, recursive application of the monotonicity assumption yields $H^n_{\mu}J^* \leq J^*$ for each $n$. Since $\rho(H^n_{\mu}J^*, J_{\mu}) \to 0$, one has $J_{\mu} \leq J^*$ for each $\mu$. Since $J_{opt}(x) = \sup\limits_{\mu\in M} J_{\mu}(x) \implies J_{opt} \leq J^*$. Hence, $J_{opt} = J^*$, under the assumptions. $\blacksquare$

Theorem 3 concludes that the solution to $J^*(x) = \sup\limits_{u_x\in U(x)} h(x,u_x,J^*)$ is unique and is $J_{opt}$, the optimal reward-to-go function. Furthermore, a policy $\mu$ is called \say{$\epsilon$-optimal} if $\rho(J_{\mu}, J_{opt}) \leq \epsilon$ and \say{optimal} if $J_{\mu} = J_{opt}$. Corollary 1 and Theorem 3 demonstrate the existence of an $\epsilon$-optimal policy, and Corollary 2 gives sufficient conditions for existence of an optimal policy.

If a sequence $\{J_n\}$, $n=0,1,\ldots$ satisfies $J_n \geq J_{n-1} \forall n$ we write $\{J_n\}\uparrow$. Lemma 1 contains useful consequences of the monotonicity assumption.

Lemma 1: 
\begin{displayquote}
Suppose the monotonicity assumption is satisfied. If $J\geq J'$, then $AJ \geq AJ'$. If $AJ'\geq J'$, then $\{A^n J'\}\uparrow$. If $H_{\mu}J' \geq J'$, then $\{H^n_{\mu}J'\}\uparrow$.
\end{displayquote}

Proof: By definition, $AJ \geq H_{\mu}J$. Suppose $J\geq J'$, then $H_{\mu}J \geq H_{\mu}J' \implies AJ \geq H_{\mu}J' \forall \mu$. Hence, $AJ \geq AJ'$. If $AJ' \geq J'$, the recursive application of the proceeding statement yields $\{A^n J'\}\uparrow$. If $H_{\mu}J' \geq J'$, then recursively applying the monotonicity assumption yields $\{H^n_{\mu}\}\uparrow$.

The last two theorems are related to an $N$-stage contraction but further coverage of these are omitted from the survey. I merely state the $N$-stage contraction assumption: 
\begin{displayquote}
For each $\mu$, the operator $H^N_{\mu}$ has modulus $c$ or less, where $c$ and $N$ are independent of $\mu$, and $c < 1$. Furthermore, for each $\mu$, $H_{\mu}$ has modulus $1$ or less. These theorems are not necessary for understanding the next paper, but could be useful for other researchers interested in such finite horizon setups.
\end{displayquote}
In his appendix, a metric is defined for comparing reward-to-go functions. Consider a set $V$. A function $\rho$ mapping $V \times V$ to $\mathbb{R}$ is a metric if
 \begin{enumerate}
   \item $\rho(J,J') \forall J,J' \in V$,
   \item $\rho(J,J') = 0 \text{ iff } J=J'$,
   \item $\rho(J,J'') \leq \rho(J,J') + \rho(J',J'')\  \forall J,\ J',\ J'' \in V$, and if $\rho(\cdot,\cdot)$ is a metric for $V$, then $V$ is a metric space.
 \end{enumerate}

Let $A: V \to V$. The function $A$ is called a contraction mapping if for some $c$ satisfying $0\leq c < 1$, it is true that $\rho(AJ,AJ') \leq c\rho(J,J') \forall\  J,J'\in V$. The element $J^*$ of $V$ is called a fixed point of $A$ if $AJ^* = J^*$. A sequence $\{J_n\}$, $n=0,1,\ldots \in V$ is a Cauchy sequence if $\forall \epsilon > 0,\  \exists N$ such that $\rho(J_m, J_n) < \epsilon \forall\  m,n > N$. A metric space is called \say{complete} if for every Cauchy sequence $\{J_n\}$, $n=0,1,\ldots \exists\  J\in V$ such that $\lim\limits_{n \to \infty} \rho(J_n, J) = 0$. For a map $A$ of $V$ into itself, the function $A^n$ is defined recursively by $A' = A$, and $A^{n+1} = A(A^n)$.

Fixed-point theorem: 
\begin{displayquote}
Let $V$ be a complete metric space. Suppose for some $N\in\mathbb{Z}^+$ that $A^N$ is a contraction mapping. Then, $A$ has a unique fixed-point $J^*$. Furthermore, $\lim\limits_{n \to \infty} \rho(A^n J, J^*) = 0 \forall J \in V$.
\end{displayquote}

\subsection{Paper 4: Sondik, ``The Optimal Control of Partially Observable Markov Processes Over the Infinite Horizon: Discounted Costs"}

\subsubsection{Overview of the paper}
 This paper, \cite{sondik1978optimal}, examines the extension of the problem of finite horizon POMDPs to the infinite horizon with discounting of future costs. The paper introduces a new type of stationary control policy, finitely transient control policies which are a subset of the stationary control policies, that leads to a piece-wise linearity property of the cost-to-go functions similar to the finite horizon case. In Sondik's words,\say{Finitely transient policies, which are discussed extensively in a later section, are a particular class of control policies for partial observable Markov processes that generate dynamics equivalent to those of a completely observable process.}. In the paper, finitely transient policies provide a convenient conceptual bridge between the completely observable and the partially observable Markov processes. Section 2 describes the optimization problem. Section 3 defines and derives several properties of finitely transient policies. Later sections use this policy to derive approximations to the stationary policy, and develop a policy iteration algorithm.



From \cite{blackwell1965discounted}, it follows that the minimum expected cost as a function of the initial information vector takes the form of the Bellman equation with $\pi$ as the state, and the Bayesian filter being the \say{dynamics} for the state. Furthermore, a measurable policy (with respect to $\pi$) exists that achieves
this minimum cost (\cite{maitra1968discounted}). That is, given any $\pi$, we know the optimal control alternative! This paper, \cite{sondik1978optimal} focuses on a second solution technique, a generalization of Howard's policy iteration algorithm for completely observable processes \cite{howard1971dynamic}. In general, for the infinite horizon, the optimal cost-to-go functions are not piece-wise linear. However, if the stationary policies are finitely transient, then, we get the piece-wise linearity property already familiar to us from the finite-horizon results. The papers, \cite{kakalik1965optimum} and \cite{satia1969markovian} have considered approximate solutions to the general problem. 


\subsubsection{Infinite Horizon POMDP model with Discounting}
Most of the same notation carries over from the paper \cite{sondik1971optimal}. Consider an $n$-state $\mathcal{X}=\{1,2,\ldots,n\}$ M.C.. The main difference in the cost-to-go interpretation is that $J_k(\cdot)$ represents the optimal total cost from stage $k$ to forever. In the paper, he focused on reward maximization. Here, we focus on cost minimization, the natural DP inclination. I use $u\in \mathcal{U}$ to denote a control alternative from the set of alternatives, respectively. $p_{ij}(u)$ are the controlled state transitions. Observation transitions from a stage are denoted by $q_{jy}(u)$ for some $y\in\mathcal{Y}=\{1,\ldots,M\}$. $Q_y(u)$ is the matrix that houses the transitions from all states to any observation. $P(u)$ houses all of the state transitions under some control $u$. In this model, control influences observation probabilities.

Now we can define $c(u)$, the expected cost vector, a column vector of length $n$. $c_i(u)$ is the expected immediate cost of transitioning from $i$ using action $u$. We define the current state information vector $\pi = [\pi(1),\ldots,\pi(n)]$, and also call it the \say{belief-state} interchangeably with the term \say{information-vector} or information state or more simply, state.

As in the paper \cite{sondik1971optimal}, the update for $\pi$ is $\pi'(j) = T(\pi|u,y) = \sum_{i}\pi(i)p_{ij}(u)q_{jy}(u)$, which is derived from an application of Baye's rule. This equation can easily be expressed in matrix-vector form as $$ \pi' = T(\pi|u,y) = \frac{\pi P(u) Q_y(u)}{\sigma_y},$$ where $\sigma_y = \pi P(u) Q_y(u) \mathbf{1}$ and $\mathbf{1}$ is an all ones column vector. From \cite{sondik1971optimal}, we know that $\pi_k$ is a M.C.. An extra assumption made in the current paper, \cite{sondik1978optimal}, is that $T(\pi|u,y)$ is a $1-1$ function in $\Pi$, the belief-space. The belief-space for any given state space size, is the space of all belief-states, which are points in this space. It is defined as: $$ \Pi = \{\pi : \sum_i \pi(i)=1,\  \pi(i)\geq 0\  \forall i\}.$$

Given discount factor $\beta, 0\leq \beta < 1$, the goal is to minimize the cost of the POMDP over all time. To achieve this goal, let's start with some basic new definitions. A finite connected partition of $\Pi$ is $[V_1,\ldots,V_l]$ which is a finite collection of mutually exclusive and exhaustive connected subsets of $\Pi$. A \say{control function} or policy $\mu: \Pi \to U$ is admissible if $\exists$ a finite connected partition of $\Pi$ such that $\mu(\pi)$ is constant over each set $V_j$. The set of such policies is denoted by $\Delta$.

Use the notation that $\mu_k(\pi_k)$ represents the alternative to be used at time $k$ if the state is $\pi_k$. Given an initial state $\pi_0$, define the discounted expected value control problem as:
$$ \min\limits_{\mu_0,\mu_1,\ldots} E_{\pi_0}\{ \sum_{k=0}^{\infty}\beta^k\pi_k c(\mu_k(\pi_k)) \},$$ where we need to minimize this expression over all possible sequences of control functions $\{\mu_0,\mu_1,\ldots,\}$. The policy is called \say{stationary} if we use the same control function at every stage. Denote it by $(\mu)^{\infty} = (\mu,\mu,\ldots)$.

The paper, \cite{blackwell1965discounted}, presented a Bellman like functional equation for the discounted POMDP with initial state $\pi$ as:
$$ C^*(\pi) = \min\limits_{a} [\pi q^a + \beta \sum_{\theta}\{\theta|\pi,a\}c^*(T(\pi|\theta,a))].$$
In updated notation, this becomes
$$ J^*(\pi) = \min\limits_{u}[\pi c(u) + \beta\sum_{y}J^*(T(\pi|u,y))\sigma_y]. $$
The paper \cite{maitra1968discounted} showed such a measurable policy exists that achieves the minimum cost. This optimal policy is in fact stationary. Denote it by $(\mu^*)^{\infty}$; then, $\mu^*(\pi)$ is the minimizer for the functional in terms of $J*(\pi)$. Hence, we can write:
$$ J^*(\pi) = \pi c(\mu^*) + \beta \sum_{y} J^*(T(\pi|\mu^*,y))\sigma_y .$$ The rest of this paper focuses on building a generalization of Howard's Policy Iteration algorithm (P.I.A) \cite{howard1971dynamic}. Define $J(\pi|\mu)$ as the expected cost of following the stationary policy $(\mu)^{\infty}\  \forall k\  \in \mathbb{Z^+}$. Theorem 1(Howard-Blackwell P.I.A): Let $\mu'(\pi)$ be the control function defined as the index $u$ minimizing $U_u[\pi, J(\cdot|\mu)], $where $U_u[\pi, J(\cdot,\mu)] = \pi c(u) + \beta\sum_{y}\sigma_y J(T(\pi|u,y)|\mu)$. Then, $J(\pi|\mu') \leq J(\pi|\mu)\  \forall \pi\in \Pi$. Furthermore, if $J(\cdot|\mu) \neq J^*$, there is some $\pi$ such that $J(\pi|\mu') < J(\pi|\mu)$.

Note it is useful to develop a measure of distance between $J^*,\  J(\pi|\mu)$. Typically, this is some kind of stopping rule based on a number of iterations of Theorem 1. The measure and Theorem 1 together form a \say{policy iteration} algorithm. Regardless, there are a few implementation issues to Theorem 1. First, one must be able to compute $J(\cdot|\mu)$, the cost of some stationary policy $\mu$. $U_{\mu}$ is defined for $\mu(\pi)$ as $$ U_{\mu}(\pi, f) = \pi c(\mu) + \beta \sum_{y}\sigma_y f(T(\pi|\mu,y))$$ is a contraction mapping, under the sup norm metric $ ||f|| = \sup\limits_{\pi\in\Pi}|f(\pi)|$. It follows from the fixed-point theorem that we can show that the sequence of functions $f^n(\cdot)$ where $f^{n+1}(\pi) = U_{\mu}(\pi,f^n), n\geq 1$ converges to $J(\cdot|\mu)$ in the sense of the sup norm. Furthermore, $J(\cdot|\mu)$ is the unique bounded solution to these iterate functions $f^n(\cdot)$. Whether or not $J(\cdot|\mu)$ has the piece-wise linear property is addressed later. If the answer is yes to the above question, one can apply the methods developed in paper \cite{sondik1971optimal}.

The second major impediment to Theorem 1 for computing $J^*(\pi), \mu^*(\pi)$ is that minimizations must be carried out for every point in the space $\Pi$.  The concave hull of $J(\cdot|\mu)$ is used to simplify Theorem 1 instead of using $J(\cdot|\mu)$ directly. This enables policy improvement with each successive iteration, thus obtaining a lower expected cost.

\subsubsection{The Expected Cost of Finitely Transient Policies}

I start this section with a few new definitions. \say{Piece-wise linear}: a real-valued function $f(\cdot)$ over $\Pi$ is termed piece-wise linear if it can be written as $f(\pi) = \pi\alpha_j\  \forall \pi\  \in V_j \subset \Pi$, where $V_1,\ldots,V_l$ is a finite connected partition of $\Pi$, and $\alpha_j$ is a constant column vector for $\pi\in V_j$.

We know from \cite{sondik1971optimal} that the finite horizon cost-to-go functions are piece-wise linear, but for the infinite horizon, $J^*(\cdot)$ need not be. However, if $\mu^*$ belongs to a class of policies termed Finitely Transient (F.T.), then the piece-wise linearity in $J^*$ over the infinite-horizon is preserved.

\begin{definition}
Finitely Transient (F.T.) policy: A stationary policy $(\mu)^{\infty}$ is F.T. iff $\exists\  n\in \mathbb{Z}^+ < \infty$ such that $D_{\mu}\cap S^n_{\mu} = \phi$, where $\phi$ is the empty set. The smallest such integer $n$ is termed the "index" of the F.T. policy and labeled $n_{\mu}$.
\end{definition}

We have to make three additional definitions to be able to understand the above definition. Def i) for $(\mu)^{\infty}$, let $T_{\mu}$ be the set function defined for any set $A \subset \Pi$ by:
$$ T_{\mu}(A) = \text{ closure }\{ T(\pi|\mu, y) : \pi\in A, \forall y\}.$$ $T_{\mu}(A)$ represents the set of all next possible belief-states. Def ii) consider the sequence $S^n_{\mu}$ defined initially as $S^0_{\mu} = \Pi$ with the rest of the sequence's elements defined by the recursion $S^n_{\mu} = T_{\mu}(S^{n-1}),\  n\geq 1$. $S^n_{\mu}$ contains all belief states that could occur at the $n$th stage of operation. The actual sequence, in practice, is influenced by $\pi_0$, the initial belief-state. Lastly, Def iii) $D_{\mu} = \textit{ closure }(\{\pi:\mu(\pi)\textit{ is discontinuous at }\pi \}).$ 

Next, a Lemma is given that relates a F.T. policy's index to the above sequence of sets. Lemma 1: If $(\mu)^{\infty}$ is F.T. with index $n_{\mu}$, then $D_{\mu}\cap S^n_{\mu} = \phi\  \forall n\geq n_{\mu}$. The proof is in the paper and uses induction to show the nesting of the sets $S^n_{\mu}$ to establish the claim. Results are given next that lead to F.T. policy's expected cost having the piece-wise linearity property. These are structural results that give insight into POMDPs and are fundamental for calculating $J(\pi|\mu)$ for arbitrary $(\mu)^{\infty}$.

\begin{definition}
Markov Partition (M.P.) as a partition $V=[V_1,V_2,\ldots]$ of $\Pi$ that possesses the following two properties with respect to a stationary policy $(\mu)^{\infty}$ is said to be a M.P.. Property a) All points in the set $V_j$ are assigned the same control by $\mu$ [i.e., if $\pi^{(1)},\pi^{(2)}\in V_j$, then $\mu(\pi^{(1)})=\mu(\pi^{(2)})$]. Property b) Under the mapping $T(\cdot|\mu,y)$, all points in $V_j$ map into the same set. 
\end{definition}

The relationship between the sets $V_j$ and the mappings $T(\cdot|\mu,y)$ is given by the Markov Mapping (M.M.) $\nu(j,y)$ such that if $\pi \in V_j$, then $T(\pi|\mu,y) \in V_{\nu(j,y)}$. If this property holds, we shall say that one set maps completely into another for each output. Define \say{Equivalence to $\mu$}: The M.M. $\nu(\cdot,\cdot)$ and the M.P. $V$ satisfying Property b) is said to be "equivalent to $\mu$". Lastly, the control, defined by $\mu_j$, is constant over each $V_j$ in the partition, i.e., if $\pi\in V_j$, then $\mu(\pi) = \mu_j$. Lemma 2 relates F.T. policies to M.P. and M.M.s. Lemma 2: For every F.T. policy $(\mu)^{\infty}, \exists$ a M.P. of $\Pi, V=[V_1,V_2,\ldots]$ and M.M. $\nu$ satisfying properties a) and b). To prove this lemma requires two more definitions. We must define a D-sequence of sets, and the other a V-sequence of partitions. i) A D sequence of sets $D^0, D^1,\ldots$ is defined as follows: $$D^0=D_{\mu}, D^n = \{\pi:T(\pi|\mu,y)\in D^{n-1}, \textit{ for some } y , n \geq 1\}.$$ ii) A V sequence of partitions of $\Pi$, $V^0,V^1,\ldots$ is defined as follows: $V^0$ is the smallest collection of mutually exclusive connected sets in $\Pi$ satisfying property a). We say that $V^0$ is induced by $\mu$, and note that the set $D^0$ forms the boundaries of the sets of $V^0$. Similarly, for $k\geq 1$, $V^k$ is the refinement of $V^{k-1}$ where the set $D^0\cup \ldots \cup D^k$ forms the set boundaries of $V^k$. Note that for each stage $k$, $V^k$'s  points get mapped into the same set of $V^0$.

The proof of Lemma 2 lies in contradicting the index of the F.T. assumption, $n_{\mu}$. Thus, he proves that every set in $V^{n_{\mu}-1}$ maps completely inside another set in $V^{n_{\mu}-1}$, but cannot be on a boundary where $\mu(\pi)$ would be discontinuous. Hence, $V^{n_{\mu}-1}$ is the required partition. Then, we can construct a M.M. easily $\nu$ such that if $\pi\in V_j$, then $T(\pi|\mu,y)\in V_{\nu(j,y)}$. Then, Lemma 3 follows next. 

Lemma 3: 
\begin{displayquote}
The policy $(\mu)^{\infty}$ is F.T. with $n_{\mu}=n$ iff $D^n$ is the first empty set in the sequence $D_{\mu}, D^1, D^2,\ldots$
\end{displayquote}

After this definition, an example illustrating a F.T. policy is given where $n_{\mu}= 5$. The figure is quite busy and near impossible to interpret. He does not give the costs for the problem so it cannot be reconstructed. The next section is an overview of the main approximation methods for POMDPs before ending with the wide range of applications. Effective approximations to POMDPs in their many settings are so important to learning \say{good} policies in a cheap way to save vast amounts of computations, while still being as close to optimal as possible should be the goal for the methods of the next section for whatever notion of optimality one is considering for the POMDP.


Next in this paper, it is shown that $J(\pi|\mu)$ is piece-wise linear when $(\mu)^{\infty}$ is F.T.. Lemma 4: Let $\mu\in \Delta$; then, $J(\pi|\mu)$ can be written as $J(\pi|\mu) = \pi\alpha(\pi|\mu)$ where $\alpha(\pi|\mu)$ is a column vector of length $n$, that is the unique bounded solution to the vector equation $$\alpha(\pi|\mu) = c(\mu(\pi)) + \beta \sum_{y}P(\mu(\pi))Q_y(\mu(\pi))\alpha[T(\pi|\mu,y)|\mu].$$ See the paper for the simple proof, but it does rely on a contraction mapping assumption which is easy to verify. Lemma5: If $(\mu)^{\infty}$ is a F.T. policy, the vector function $\alpha(\pi|\mu)$ assumes only a finite number of values over $\Pi$; furthermore, these values are related by a set of linear equations. Proof: Let $(\mu)^{\infty}$ be a F.T. policy. From Lemma 2, we can partition $\Pi$ into $V_1,...,V_m$ with properties a) and b). Now, if we define $\alpha(\pi|\mu)=\alpha_j$ for each $\pi\in V_j$, and select any set of $m$ points $\pi^i\in V_i, 1\leq i\leq m$, then
$$ \alpha(\pi|\mu) = c(\mu(\pi)) + \beta \sum_{y}P(\mu(\pi))Q_y(\mu(\pi))\alpha[T(\pi|\mu,y)|\mu],$$ becomes the following set of vector equations:
$$ \alpha_i = c(\mu(\pi^i)) + \beta\sum_{y}P(\mu(\pi^i))Q_y(\mu(\pi^i))\alpha_{\nu(i,y)}\  \forall 1\leq i\leq m. $$ Recalling that $\mu_i = \mu(\pi), \pi\in V_i$ we have:
$$\alpha_i = c(\mu_i) + \beta\sum_{y}P(\mu_i)Q_y(\mu_i)\alpha_{\nu(i,y)}. $$ This set of linear equations can be solved uniquely for $\alpha_1,\alpha_2,\ldots,\alpha_m$. A simple example is given of Lemma 5 where $\mu(\pi)$ is constant, and we assume that $\alpha(\pi|\mu) = \alpha\  \forall \pi$. Then, substitution into the original equation of $J(\pi|\mu)$ yields $\alpha = c + \beta P\alpha$ which is referred most often in the literature as the discounted Poisson equation; $\alpha$ is our optimal cost-to-go function and is obtained by $$\alpha = [I-\beta P]^{-1}c.$$ Thus, $J(\pi|\mu) = \pi\alpha$; further, note that $\alpha_i$ is the support function of $J(\pi|\mu)$; $J(\pi|\mu)$ has the same gradient for all $\pi\in V_i$.

Based on this analysis, I think I have a new interesting idea for developing a new type of policy approximation. Perhaps we can develop $\alpha$ vectors associated with each observation $y$ as
$$ \alpha_y = [I-\beta P(u)Q_y(u)]^{-1}c(u)$$ and possibly minimize each of these vectors $\alpha_y$ for each $y$ over all $u$. I would like to call this equation, the Observational Poisson Equation (OPE). Hence, we would only need to consider $\mathcal{Y}^{\mathcal{U}}$ number of policies instead of $\mathcal{X}^{\mathcal{U}}$ number of policies which in most problems I have seen is a much larger class of policies based on the cardinialities of the sets. Along these lines, we could also develop a new object associated with the observations as the expected total cost-to-go from this observation value $y$, and relate this observation to a cost-to-go by developing a value iteration routine using the current belief-state also.

Continuing with the paper,\newline
Theorem 2: If a policy $(\mu)^{\infty}$ is F.T., then $J(\pi|\mu)$ is piece-wise linear. Proof: follows directly from Lemmas 4 and 5. The theorems will be be presented without proof for the rest of this paper. The next section deals explicitly with the approximation problem of $J(\pi|\mu)$ by different methods. 


\subsubsection{The Approximation of $J(\pi|\mu)$}

Since not all policies are F.T., their costs cannot be computed as compactly as linear vector equations. He presents a bound on the true stationary cost-to-go function $J(\pi|\mu)$ and it's piece-wise linear approximation. Denote the approximation mapping as $\hat{\nu}$. From before, we constructed the partition $V$ of $\Pi$ from the sets $D_{\mu}, D^1, D^2,\ldots$, and used these as boundaries of the sets $V_j$ in the partition. Suppose for some integer $k$ we have $D_{\mu}, D^1, D^2,\ldots, D^k$ for policy $(\mu)^{\infty}$. If $D^k = \phi$, then from Lemma 3, $(\mu)^{\infty}$ is F.T. with degree $n_{\mu} \leq k$. If $D^k=\phi$, then we do not know yet whether $(\mu^{\infty})$ is F.T. Let the boundaries of the sets in the partition of $\Pi, V^k = [V_j]^k$ be formed by the set $D_{\mu}\cup D^1 \cup \ldots \cup D^k$. Using $V^k$, we can construct an approximation mapping $\hat{\nu}$ that is used to approximate $(\mu)^{\infty}$. Since $\hat{\nu}$ is constructed from $V^k$, $k$ is called the degree of the approximation. We can arbitrarily select any point $\pi^j$ in each set $V^k_j$. Then, the mapping $\hat{\nu}$ is defined as follows: $$\text{ if }T(\pi^j|\mu,y)\in V^k_l \textit{ then } \hat{\nu}(j,y) = l.$$ The mapping $\hat{\nu}$ can be used to build a piece-wise linear approximation to $J(\pi|\mu)$, which is shown later to produce a bound on the error proportional to $\beta^k$. This approximation to $J(\pi|\mu)$, denoted as $\hat{J}(\pi|\mu)$ is defined as $\hat{J}(\pi|\mu) = \pi \hat{\alpha}_j,\  \pi\in V^k_j$, where the vectors $\hat{\alpha}_j$ are chosen to satisfy the set of linear equations:
$$\hat{\alpha}_j = c(\mu_j) + \beta \sum_{y}P(\mu_j)Q_y(\mu_j)\hat{\alpha}_{\hat{\nu}(j,y)},$$ where $\mu_i = \mu(\pi)$ for $\pi \in V^k_i.$ An error bound on the difference between $J(\pi|\mu),\  \hat{J}(\pi|\mu)$ is given next. Theorem 3: If $\hat{\nu}$ is constructed from a partition $V^k$ of degree $k$, then $$ ||J(\cdot|\mu)-\hat{J}(\cdot|\mu)||\leq \frac{\beta^k}{(1-\beta^k)}\frac{K}{(1-\beta)}, $$ where $K= \max\limits_{u,i}[c(u)]_i - \min\limits_{u,i}[c(u)]_i$. See the paper for the proof.

Section 5 introduces a generalized policy iteration (P.I.) algorithm to improve the stationary policy. It uses the concave hull of $J(\pi|\mu)$ in the P.I.. But first some more definitions are needed. First, from \cite{stratonovich1965conditional}, a class of policies called Semi-Markov are introduced. A Semi-Markov policy is a sequence of functions $(\psi^1,\psi^2,\ldots)$, where $\psi^t: \Pi \times \Pi \to U$. $\psi^t(\pi^0,\pi)$ is a control to be used at the $t$-th time period of system operation if the state at time $t$ is $\pi$ and the initial state was $\pi^0$. It is shown in this section that the concave hull of $J(\pi|\mu)$ is the cost function of a Semi-Markov policy and thus can be used in P.I..

The \say{Concave Hull} of $J(\pi|\mu)$ is denoted as $\bar{J}(\pi|\mu)$ and defined as
$$\bar{J}(\pi|\mu) = \min\limits_{\pi'} \pi \alpha(\pi'|\mu), $$ where $\alpha(\pi|\mu)$ satisfies $$ \alpha(\pi|\mu) = r(\mu(\pi)) + \beta \sum_{y} P(\mu(\pi)) Q_y(\mu(\pi))\alpha[T(\pi|\mu,y)|\mu].$$ Note that the $\alpha(\pi|\mu)$s are essentially the support hyper-planes of $J(\pi|\mu)$. Consequently, $\bar{J}(\pi|\mu)$ are formed from these hyper-planes.

The rest of the paper assumes that $J(\pi|\mu)$ is piece-wise linear and consists of a finite number of linear segments that provide sufficient structure to develop the P.I. algorithm. Theorem 4: If $\bar{J}(\pi|\mu)$ is substituted for $J(\pi|\mu)$ in the functional from before, then $J(\pi|\mu')\leq J(\pi|\mu)\  \forall \pi\in \Pi$, where $\mu'$ is defined in Theorem 1. Furthermore, read through Theorem 6 up to the P.I.A. section. Sufficient understanding of Paper $\#2$ allows understanding of this material. I will present the lemmas and theorems.

Lemma 6: The expected cost of following the semi-Markov Policy based on the $k$th degree approximation mapping $\hat{\nu}$ is $\bar{\hat{J}}(\cdot|\mu)$. Then, it is straight forward to see that $\bar{\hat{J}}$ can be used instead of $J(\cdot|\mu)$ in the older P.I.A algorithm of Howard. Theorem 5: If $\bar{\hat{J}}(\cdot|\mu)$ is used in P.I. in place of $J(\cdot|\mu)$, then $J(\cdot|\mu')$, the expected cost of the resultant stationary policy $(\mu')^{\infty}$ satisfies for all $\pi \in \Pi:\  J(\pi|\mu')\leq \bar{\hat{J}}(\pi|\mu) \leq \hat{J}(\pi|\mu) \leq J(\pi|\mu) + \epsilon_k$, where $\epsilon_k$ depends on the degree of the approximation to $(\mu)^{\infty}$ and $\epsilon_k \to 0$.

Next, a bound on the distance between $\bar{\hat{J}}$ from $J^*$ is presented. Theorem 6: $||\bar{\hat{J}}(\cdot|\mu) - J^*(\cdot)|| \leq (1-\beta)^{-1}||\bar{\hat{J}}(\cdot|\mu) - \min\limits_{u}U_u[\cdot,\bar{\hat{J}}(\cdot|\mu)]$, where $U_u$ is defined in Theorem 1.

\subsubsection{Discussion of the Generalized Policy Iteration Algorithm}
The basic idea is that the algorithm iterates through a succession of approximations to stationary policies using the expected cost of each approximation policy as a basis for P.I.A.. This is different from P.I. implied by Theorem 1 because the new P.I.A uses the more easily computed expected costs of approximation to stationary policies, instead of the true possibly nonlinear cost functions of the stationary policy $(\mu)^{\infty}$. He provides a logical schematic block-diagram of how the algorithm should operate.

Theorem 7: The policy iteration algorithm defined in figure 3 converges to a policy arbitrarily close to the optimal policy. See paper for the proof and steps to run the algorithm using the approximate M.M. $\hat{\nu}$ and the computed partitions $V^k$. Closing remarks: He says his new P.I.A. is similar to grid approximation of $\Pi$ like \cite{eckles1968optimum} and \cite{kakalik1965optimum}. For the algorithm, the partition $V^k$ can be considered to represent a set of belief states, and the vectors $\hat{\alpha}_j$ each represent a vector cost over the states. In the next section I begin with the fascinating subject of approximation methods for POMDPs that aims to simplify the computation of the optimal cost-to-go functions and the optimal-policies and is the main topic of this survey paper.

\section{Approximation Techniques for Solving POMDPs}
\label{approximations}

A solution in terms of a policy for a POMDP is in general a mapping from belief space to control space. In the notation I have used throughout the survey, that analytically means $u=\mu(\pi)$. An optimal policy, for the infinite horizon is characterized by a cost-to-go function for minimization of costs as $J_\mu(\pi_0)= E[\sum_{k=0}^{\infty} \gamma^k c(\pi_k,\mu(\pi_k))]$ for some stationary policy $\mu$. In general, one could insert $\mu_k$ in place of $\mu$ for a non-stationary policy. Even for a finite horizon, the POMDP has proven to be P-SPACE complete. For this reason, most applications rely on a simplified MDP-based approximation. These reasons again show how complex it is to develop an exact solution to a POMDP. This motivates us to develop the best approximation techniques that are possible. First I cover some techniques from papers that deal with exact methodology, and then I review papers from one category of approximation already mentioned: reduction to MDP from POMDP. Within the category of exact methods, I cover a paper that uses an exact value iteration procedure for POMDPs.

This section is all about presenting more computationally efficient algorithms for finding POMDPs optimal policies. I start with the paper \cite{cassandra1994acting} called the Witness algorithm. However, the details of this algorithm are left to a separate technical report. Some results are given from this algorithm and a few simple toy models are given. First, I would like to recast his approach into cost minimization from reward maximization and continue with our notation already developed. No new notation is required for reading and understanding this paper. In this paper, an interesting block-diagram illustrates the dynamics of a POMDP. Below it is reproduced from the paper \cite{cassandra1994acting}:

\begin{figure}[hbt!]
  \centering
{\includegraphics[height=4cm, width=10cm]{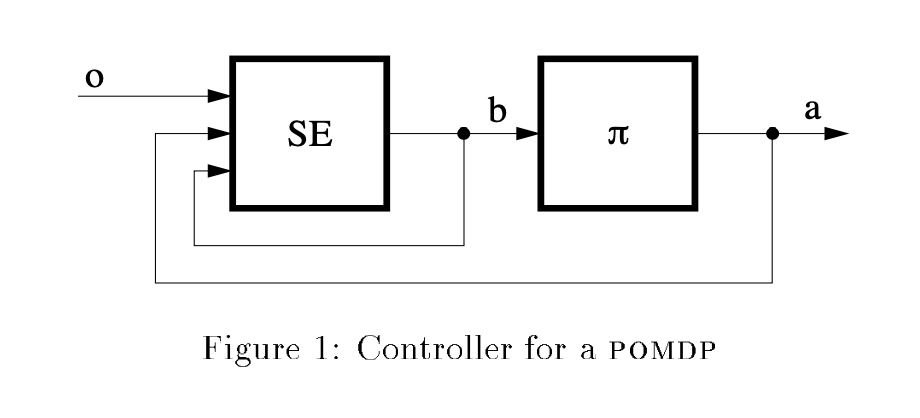}}
    \caption{State Estimator and Controller Block Diagram}
  \label{fig:SE}
\end{figure}

The dynamics show $o\equiv y$ an observation, $\mathbf{b}$ a belief-state, and $a\equiv u$, a control being input into the \say{S.E.} block, which stands for state estimator. The block \say{S.E.} produces a new belief state at its output. This is the optimal Bayesian filter or belief update we have studied already.

For POMDPs, since the states are not completely observable, we can only learn about them through the observation process. Given a finite set of observations $\mathcal{Y} = \{1,2,\ldots,m\}$, we have a discrete pmf over $\mathcal{Y}$ given by $Pr(Y=y|X=x, U=u)$, which is the probability of observing $y$ given the true current state is $x$ and we choose $u$ in the previous period. Another common notation for belief-state is $\mathbf{b}$ instead of $\mathbf{\pi}$. The vector $\mathbf{b}$ models the internal belief of the true state $x$ as the component $b(x)$. The other component in the figure is the policy which maps $b \to \mathcal{U}=\{1,2,\ldots,l\}$, where $\mathcal{U}$ is the finite set of controls which is the same as in our basic universal model of POMDPs.

The formula for $b'(x')$, i.e., the belief of state $x'$ at the next stage is:
$$ b'(x') = \frac{q_{x'y}(u)\sum\limits_{x}P_u(x,x')b(x)}{\sum\limits_{x'}q_{x'y}(u)\sum\limits_{x}P_u(x,x')b(x)} ,$$ is our formula for SE given our transition models, observation transition probabilities, and the previous belief-state.


An example of a type of reduction approach is based on using the belief state with a $Q$-function based on the $Q$-learning algorithm of Watkins and Dayan is denoted as $Q^*_{CO}$. Here, CO stands for completely observable. This would be the optimal state-control cost for us based on the completely observable MDP $(\mathcal{X},\mathcal{U},P_u(x,x'),c(x,u),\gamma)$. Given a belief-state $\mathbf{b}$, generate the control $u$ s.t.  $$\arg\min\limits_{u\in \mathcal{U}} \sum_{x}b(x)Q^*_{CO}(x,u). $$ This pretends that the stochasticity is only present for the immediate stage and that the environment is CO for all future stages. The main issue with this type of reduction is that this kind of approximation can lead to policies that do not explore much or seek to gain more information.

The next paper I review by Chrisman \cite{chrisman1992reinforcement} helps to explain ways of overcoming these observational issues. Continuing with \cite{cassandra1994acting}, the already proven way to find optimal policies is to recast the POMDP as a continuous state MDP with $\mathbf{b}$ as the state. For any action $u$ and belief $\mathbf{b}$, there are only $|Y|=m$ successor beliefs $\mathbf{b}'$ that are possible. With this in mind, define a new belief-state transition function as $$ \tau(\mathbf{b},u,\mathbf{b}') = \sum\limits_{\{y\in\mathcal{Y}:SE(\mathbf{b},u,y)=\mathbf{b}'\}}\operatorname{Pr}(y|u,\mathbf{b}).$$ Define the cost function as $$ \rho(\mathbf{b},u) = \sum\limits_{x\in\mathcal{X}}b(x)c(x,u) .$$ This allows the agent to minimize costs using its current belief-state. The agent cannot fool itself by the design of S.E.

Next, the method of value iteration for discounted infinite horizon for POMDPs is presented. The optimal cost-to-go $J^*(\mathbf{b})$ from belief-state $\mathbf{b}$ , analogous to the finite-state MDP Bellman equation, represents an infinite expected sum of discounted costs from the original belief-state $\mathbf{b}$, with the optimal policy being executed for all subsequent stages. The form of $J^*(\mathbf{b})$ is as follows:


$$ J^*(\mathbf{b}) = \min\limits_{u\in U}\biggl[\rho(\mathbf{b},u) + \gamma \sum_{\mathbf{b}'\in B}\tau(\mathbf{b},u,\mathbf{b}')J^*(\mathbf{b}')\biggr].$$

One could consider value iteration for a finite horizon, but it is known that in the limit that $J^*_k(b) \to J^*(b)$ approaches the infinite-horizon cost-to-go function, where $k$ is the number of stages. Then, VIA is presented as:


\begin{footnotesize}
\begin{algorithm}[H]
\KwResult{Converged cost-to-go function within $\epsilon$ optimality}
 Let ~ $J_0(\mathbf{b})=0\  \forall \mathbf{b}\in B$\;
  \textbf{Do}
  {
    Let\  $k += 1$, $\forall$ {$\mathbf{b}\in B$}\;
      { $~$ $J_k(\mathbf{b}) = \max\limits_{u\in U}[\rho(\mathbf{b},u) + \gamma \sum\limits_{\mathbf{b}'\in B}\tau(\mathbf{b},u,\mathbf{b}')J_{k-1}(\mathbf{b}')$\;
    }\textbf{While}$|J_k(\mathbf{b})-J_{k-1}(\mathbf{b})| < \epsilon$
 \newline
 }
 \caption{VIA for POMDPs}
\end{algorithm}
\end{footnotesize}

He references Lovejoy, \cite{lovejoy1991survey} for more on this value iteration technique.
The cost-to-go function converges in a finite number of steps to a near optimal policy that yields a value function that is within $\frac{2\gamma\epsilon}{(1-\gamma)}$ of the optimal policy. There is one other interesting idea in \cite{cassandra1994acting}. Policy graphs are introduced as a way to visually represent policies. If we let $\nu_k$ be a set of $|\mathcal{X}|$-dimensional vectors of real numbers, the optimal $k$-stage cost-to-go functions can be written as $$J_k(\mathbf{b}) = \max\limits_{\mathbf{\alpha}\in \nu_k} \mathbf{b}\cdot \mathbf{\alpha}.$$ The policy graph succinctly describes the final optimal policy based on the final partitioning of the belief-space. For more on the policy graph idea and how it represents the optimal policy see the paper \cite{cassandra1994acting}.





In \cite{cassandra2013incremental}, Cassandra presents an algorithm that is better than the Witness algorithm by a constant factor, \cite{krishnamurthy2016partially}. An interesting result that comes from running value iteration on POMDPs in the paper \cite{cassandra2013incremental}, in terms of accuracy of the value function is the following: $$ |V_N(\pi)-V(\pi)| \leq \frac{\gamma^{N+1}}{1-\gamma}\max\limits_{x,u}|c(x,u)|, $$ where the error is based on having run value iteration for $N$ stages, a discount factor $\gamma$ and the maximum value of the cost table. See section $7.6.3$ of \cite{krishnamurthy2016partially} for more details. The rest of the focus of this section is on approximation methods of POMDPs, mainly via reduction to MDPs. The objective is to review other highly accurate, computationally friendly algorithms that can achieve near-optimal policies while still providing plenty of insight into the POMDP problem, and see what POMDP approximation methods may be extended to the multi-agent control, partially observed setting. 





Ref. \cite{chrisman1992reinforcement} addresses an important issue for both POMDPs and RL, that of world states being indistinguishable from others because of limited sensing capabilities. According to \cite{chrisman1992reinforcement}, \say{Perceptual aliasing occurs when multiple situations that are indistinguishable from immediate perceptual input require different responses from the system}. In other words, based on the observations of the system alone, it is insufficient to learn the optimal policy. The paper introduces a new approach called the \say{predictive distinctions approach}
to alleviate the effects of perceptual aliasing caused by inadequate disambiguation of the states of the system. A predictive model is developed to better account for states not directly observed. 




Ref. \cite{chrisman1992reinforcement} addresses the perceptual distinction issue in two separate ways. It also handles stochastic controls and noisy sensor observations gracefully. Lastly, it can accommodate situations where memory of the system is important. The predictive distinction approach has at its core a predictive model inside the system. The predictive model learns a transfer function from the internal state of the predictive model to control evaluation, through a Q-learning type algorithm. Note that because of perceptual aliasing, identical observations are able to be registered from distinct world states. Hence, learning the model within this framework involves learning the state transitions and observation transition probabilities, but also discovering what the states of the world are. Hence, a simplified model would take what's developed and create a model-based version if those system parameters were available.



The core idea behind their approach is that the information required for maximizing predictive ability is most often the same information missing from perceptually aliased inputs. It is assumed that at any stage $k$, the world state $x_k\in \{1,2,\ldots,n\}$. He calls each state $x_k$ as one of a finite number of classes. And from class identity we can determine the state and observation transition probabilities.

The agent also has a finite set of controls $U$. For each $u\in U,$and a pair of classes $i,j$, $p_{ij}(u)$ specifies the probability that executing action $u$ from class $i$ will move the world state to class $j$. Denote $q_y(j)$ as the probability of observing output $y$ when the world state is in class $j$. Together, $p_{ij}(u), q_y(j)$ form the complete predictive model. The belief is maintained as before in other papers as a distribution over the states of the system. Denote it as $\mathbf{\pi}_k = [\pi_k(1),\ldots,\pi_k(n)]$ for an n-state system at stage $k$. So, $\pi_k(i)$ denotes the probability that the class at time $k$ is $i$. We update the belief from one stage to the next using the optimal Bayesian filter as discussed previously. Next, I discuss the RL aspects of the paper.

Now that a predictive model is defined as part of the system, the task of the RL agent is to learn the Q-values of the state-control pairs. The system in this paper learns a function of the form $Q(\mathbf{\pi},u): \mathbb{R}^n\times U \to \mathbb{R}$, which has as arguments the belief-state and the control that is applied. To learn this Q-function, a variation on the tabular Q learning function must be applied since the belief-state is continuous valued. The modified Q learning algorithm uses a simple approximation that is later applied in their experimental problem. For every class $i$ in the predictive model, and for each control $u$, a value function $V(i,u)$ is learned. Next, we approximate $$ Q(\mathbf{\pi},u) \approx \sum_{i=1}^n \pi(i)V(i,u).$$

Using this approximation, it is not hard to imagine a multi-agent control extension with partial observability involved. In the final section of my survey paper, I discuss some papers that have applied some RL techniques to the multi-agent control problem with partial observability. This approximation is accurate when the predictive model can effectively partition classes to determine the optimal policy. This approximation is not great for performing exploration or information gathering.  Learning $Q$ now involves adjusting the values of $V$, which has the desirable property of the belief-state being used to update all entries of the value function at every iteration of the algorithm by a specific fractional value.

Learning the $Q$-function is done by adjusting $V(i,u)$:
$$ V(i,u) = (1-\beta\pi_k(i))V(i,u) + \beta\pi_k(i)(r + \gamma\nu(\pi_{k+1})),$$ where $\nu(\pi_{k+1}) = \max\limits_{u\in U} Q(\pi_{k+1},u)$. Here $r$ is the immediate reward received. A cost function $c$ could be used instead for cost minimization. The learning rate is $\beta$, and the discount factor is $\gamma$. All entries of $V(i,u)$ are updated for all $i$ after some control $u$ is selected. Conventional $Q$-learning is obtained if the belief-state is $1$ for a single component of the vector $\mathbf{\pi}_k$. At each stage of learning, the RL agent has $\mathbf{\pi}_k$ from the predictive model, and executes the control according to the largest $Q$-value $Q(\mathbf{\pi}_k,u)$. Once the observation $y$ is received, a new belief-state $\mathbf{\pi}_{k+1}$ is computed. Then, the process repeats at the next stage of learning.

Next, in the paper he addresses model learning, where based on the experience, the goal of this component of the system is to obtain the best predictive model, i.e., the most accurate $p_{ij}(u)$ and $q_{jy}(u)$ must be obtained through proper adjustments. Secondly, additional states must be added to the model, if additional distinctions are not accounted for by the model, i.e., the number of classes of the model is increased to help mitigate perceptual aliasing. It is necessary to have collected a corpus of training experience samples because of the statistical nature of this problem. In \cite{chrisman1992reinforcement}, $m$ cycles of experience are collected, recording each control-observation pair and continuously performing the modified $Q$-learning approach presented above. For this splitting problem or distinctions, this concept may be able to transfer nicely to my most recent research on multi-parameter parameterized policies \cite{bowyer2021improving}. In \cite{bowyer2021improving}, a set number of policy parameters are divided among the state space, but it would be much more natural to learn the correct number of parameters or distinction parameters for the state space as the number of states are learned in \cite{chrisman1992reinforcement}. Furthermore the research in \cite{bowyer2021improving} may be able to be used in conjunction with POMDPs and provide a bridge for working with continuous state, control and observation spaces. 

Next, the agent calls the model-learning algorithm, using the collected experience as input to produce the more powerful predictive model. These problems are broken down into sections of probability adjustment and discovering new distinctions. Consider first the probability adjustment problem. This task involves improving the model by adjusting the state-transition probabilities $p_{ij}(u)$ and observation probabilities $q_{jy}(u)$, to maximize the predictiveness of the model. Since the true states are hidden, one cannot directly count the transitions directly to build the model but must indirectly use the belief-state $\mathbf{\pi}_k$ to access the expected transitions between states. This count, in a generic sense, for some $u\in U$, if $\pi_k(i) = a,~\pi_{k+1}(j) = b$, then the expected count is $ab$. After counting all possible expected counts, divide by the total, to obtain the expected frequencies, which is an effectively updated model.

The author uses the term $\it{quiescence}$ to describe effective parameter convergence for the model. It could be set to 0.01 e.g., when no parameters change by more than this value, quiescence is reached. Next up, consider the process for discovering new distinctions, which is deciding when to add new states to the state space. This is achieved by performing two experiments under somewhat different conditions and comparing the experience data. If the difference in results is found to be significant, then it is inferred that an influence is missing from the model, and so a new distinction(state) is introduced to the model. The author remarks that this is equivalent to detecting when the Markov property of the model does not hold.



For each class $i$ of the model, the expected frequencies from each group are tallied. For each action $u$, and each class $j$, this yields two estimates $t^1_{i,j}(u),t^2_{i,j}(u)$ for the number of expected transitions from state $i$ to state $j$. The author says, the Chi-Squared test is a good method to use for this test. Similarly for the observation counts $v^1_{jy}(u), v^2_{jy}(u)$, i.e., the expected number of times $y$ is observed from class $j$. If either test exhibits a statistically significant difference in distribution, class $i$ is split into two causing the state space to grow by one extra state. Whenever, this occurs, the complete model learning algorithm must be run again.

In their experimental section they test on a docking application with two space stations and one agent that moves between them. They report positive results. For the Chi-Squared test, a significance level of $\alpha=0.05$ was used, and a random action policy was used with $\epsilon=0.1$. Despite success in learning an optimal policy, issues with exploration exist for this method. Sometimes the agent cannot tell the difference between the unexplored portions of the world and heavily explored portions of the world, i.e., until it knows when two state regions look the same from the agent's eyes. One other major drawback in their algorithm is the \say{oversplitting} concept, where their system learns many extra states than is necessary to learn the optimal policy. This leads to interest in state conglomeration strategies. He concludes with the Utile-Distinction Conjecture: \say{Is it possible to only introduce those class distinctions that impact utility? If the color of a block is perceivable but irrelevant to the agent's task, is it possible for the agent to avoid introducing the color distinction into its model, while at the same time learning distinctions that are utile?}.

The next approximation paper \cite{parr1995approximating} I review is different in its approach to the previously formulated problems. Stuart Russel develops a technique called Smooth Partially Observable Value Approximation(SPOVA), that approximates a value function $V$ using the exact belief state. The paper may lend itself useful for later combination with other approximation algorithms for attacking the multi-agent control problem with different approximation schemes, potentially for an empirical study of multi-agent POMDP approximation methods. This paper is different from other approximation methods in that it considers a differential representation to the value function. The paper starts off by reiterating the point of AI is to make optimal decisions under uncertainty. SPOVA is shown to yield better approximations to the value function over time. In the final section of the paper, SPOVA is combined with RL algorithms, which resulted in positive results for their test cases on grid world problems. 





Now, I discuss his differentiable approximation. Sondik showed that an optimal finite horizon value function can be represented as a max over a finite set of linear functions of the belief-state. For any belief-state, the value function can be represented as $$V(\pi) = \max\limits_{\gamma\in \Gamma} \pi \cdot \gamma, $$ where $\Gamma$ is the set of alpha vectors (See \ref{Exact_Methods}). Each $\gamma$ is a hyper-plane in the value space and the max of these functions forms the convex piece-wise linear surface. This representation does not extend to an infinite horizon setting for POMDPs. For an infinite horizon, he comments that since the value function may be comprised of an infinite number of pieces, it is probable that the value function is smooth in some parts of the belief space. This intuition lead Russel to reason that a convex function that behaves like max would be a suitable approximation for the infinite horizon setting. The approximation algorithm is the following:
\begin{equation}
\label{Spova}
V(\pi) = (\sum\limits_{\gamma\in\Gamma}(\pi\cdot\gamma)^k)^{\frac{1}{k}}.
\end{equation}
To simplify matters, it is assumed $V^* > 0$ and the components of the $\gamma$'s are all positive too. Since the components of the gammas are all positive, the second partial derivative of \ref{Spova} is always positive and hence the value function is convex. Good approximations to the value function or cost-to-go functions for our purposes can be achieved with low values for $k$, e.g. $k=2$. The shape of the approximation matters more here when setting $k$ than the height of the function. This algorithm is amenable to gradient descent, which is utilized often in algorithms within the ML community to adjust parameters of any given approximation function.

To devise a strategy for adjusting his approximation in the correct direction of the optimal value function, he first makes a few definitions. Define the Bellman residual error as $$ E(\pi) = V(\pi) - (R(\pi) + \beta\max\limits_{u\in\mathcal{U}} \sum\limits_{\pi'} Pr(\pi'|\pi,u)V(\pi')), $$ where we sum over next belief-states for each possible action. Then, one just adjusts the gammas in the direction that minimizes this error. Using the smooth approximation, by gradient descent and having $w$, the weight corresponds to the $j$th component of the $i$th $\gamma$ vector, the update rule for each weight is $$\gamma_{i_{j}} += \frac{\alpha E(\pi)\pi_j(\pi\cdot \gamma_i)^k}{V(\pi)^k}.$$ An appealing property of this update formula is that the $(\pi\cdot \gamma_i)^k$ factor increases as $\gamma_i$ increases, and this is modulated further by the belief-state component for the particular state $j$, here. The parameter $k$ is viewed as a rigidity parameter. When it is low the updates are larger, and when $k$ is high the updates are smaller.

See the SPOVA algorithm in Figure 2 for the complete algorithm description. Since it is impossible to visit or sample every belief-state, the author adopted a random sampling strategy for the belief-states from the belief-space. An issue was determining a proper stopping condition for the algorithm, but some suggestions were either to use a fixed number of iterations or continue collecting samples until $E$ falls below some threshold. Other notes about the algorithm are the following. There is no convergence proof, but given the optimal value function admits a PWLC representation, and enough iterations and a large enough set of alpha vectors, it is thought the SPOVA converges to the optimal value function. Russel says it would be nice to have a condition that said what is a sufficient number of alpha vectors to guarantee convergence to the correct value function in the limit.

For the 4x4 grid world studied, with the compass directions as controls, SPOVA approximation only required a single vector to learn a near optimal policy, which makes the value function linear. This was most likely possible because of the fact that beliefs that are higher near the rewarding state will lead to the correct policy in terms of optimality. For their example that means the agent must go down and to the right until he reaches that best state. The goal state is in the bottom right cell location. Later, in this survey I overview another algorithm called Linear $Q$-learning that is perfect for learning the optimal policy in this type of environment for the stated objective. The last contribution of this paper is an updated SPOVA algorithm, called SPOVA-RL that uses the model parameters to simulate transitions in the environment. This enables the agent to explore the belief-state space for finding high utility regions for performing the updates to the value function approximation. Bellman error is used again, but now it is computed with respect to the belief-state encountered during the course of the simulation  rather than a max over all possible next belief-states. For $\pi$ at time $k$ and $\pi'$ at time $k'$, the RL Bellman error is $$E_{RL}(\pi) = V(\pi) - (R(\pi)+V(\pi')).$$
The main takeaway from this algorithm is that it is more effective at exploring the belief space by more effective updating of the gamma or \say{alpha} vectors to learn a near-optimal policy quickly. In relation to other work he makes an important observation about the application of Q-learning to POMDPs. For Q-learning a model is not needed at all. The agent learns directly from experience. This is the model-free property, but it does not carry over to the POMDP setting. This is because the agents in POMDPs try to maintain a compact representation of the policy and that can only be achieved if some model is given or assumed. The previously studied paper overcame this limitation by discovering when states needed to be added to the model. Based on Russel's other comments, Q-learning for POMDPs appears to be an open research direction, certainly so for the multi-agent control setting.
 
In conclusion, the SPOVA is an approximate method for POMDPs based on a continuous and differentiable representation of the value function. Random sampling is used within a value iteration scheme, but a large number of belief states are required to be sampled. The second algorithm cures this need for a large number of belief states by performing updates in only the actually reachable belief-states, thus saving computation and speeding up convergence of the approximation to the optimal value function.

A fruitful idea for further research could be to define a $Q$-POMDP function based on the observations and controls instead of the regular $Q$-function idea of evaluating state-control pairs. Instead, one would need to define observation-control pairs. A common drawback is too much could be assumed about the model apriori, like the exact number of states. The biggest detractor is the computational complexity with the state estimation, i.e., updating the belief state. Additionally, with most previous work a lot of computational effort is wasted on control calculations because the state estimator may take awhile to become reliably accurate enough. At least one paper has taken up the challenge \cite{singh1994learning}, where they consider a useful class of non-Markov decision Processes(N-MDPs), namely POMDPs of a specific class to learn value functions of this form. The major advancement contained in this paper \cite{singh1994learning} is that they are able to learn effective policies without using a state estimator, by extending the class of policies to include stochastic policies. 
 

 



The problem is formulated as assuming some set of true world states $X = \{x_1,x_2,\ldots,x_N\}$, and a finite set of controls $U$, transition probabilities $P_{xx'}(u)$, and immediate reward $R(x,u)$ for applying control $u$ when in some state $x$. The set of observations is denoted $Y = \{y_1,y_2,\ldots,y_M\}$, where $M>0$, and the observation probabilities $P(y|x)$ are fixed for all time. The POMDP class they consider is one where the observations are labels for disjoint partitions of $X$. Next, some facts are presented using pictorial and simple counter arguments: 
 \begin{enumerate}
     \item \say{Just confounding two states of an MDP can lead to an arbitrarily high absolute loss in the return or cumulative infinite-horizon discounted payoff.}
     \item \say{In a POMDP the best stationary stochastic policy can be arbitrarily better than the best stationary deterministic policy.}
     \item \say{The best stationary stochastic policy in a POMDP can be arbitrarily worse than the optimal policy in the underlying MDP.}
     \item \say{In POMDPs the optimal policies can be non-stationary.}
     \item \say{In the class of POMDPs defined in Section 3, there need not be a stationary policy that maximizes the value of each observation simultaneously.}
     \item \say{In the class of POMDPs defined in Section 3, there need not be a stationary policy that maximizes the value of each state in the underlying MDP simultaneously.}
 \end{enumerate}
 
Under a stochastic policy $\bar{\mu}$,
The value of a state $x$, can be written as:
$$ V_{\bar{\mu}}(x) = \sum\limits_{u\in U}Pr(u|\bar{\mu},x)[R(x,u) + \gamma \sum\limits_{x'\in X}P_{xx'}(u)V_{\bar{\mu}}(x')]. $$

Next, he defines the value of an observation:
$$ V_{\bar{\mu}}(y) = \sum\limits_{x\in X}P_{\bar{\mu}}(x|y)V_{\bar{\mu}}(x),$$ where $P_{\bar{\mu}}(x|y)$ is called the asymptotic occupancy probability distribution, which can be computed from $P(y|x),~P_{\bar{\mu}}(x)$ using Bayes' rule, where $P_{\bar{\mu}}$ is the steady state distribution. The authors take the standard TD($0$) algorithm, which is used to evaluate a given policy, to now evaluate the value of an observation via a TD($0$) like algorithm:



$$V_{k+1}(Y_k) = (1-\alpha(Y_k))V_k(Y_k)+\alpha(Y_k)(R_k+\gamma V_k(Y_{k+1})).$$


To make analysis possible, they consider, \say{a fixed stationary persistent excitation learning policy, i.e., a policy that assigns a non-zero probability to every action in every state}. They define a $Q$-value like quantity for POMDPs with stochastic control $\bar{\mu}'(y)$ as:

$$Q_{\bar{\mu}}(y,\bar{\mu}') = \sum\limits_{x\in X}P_{\bar{\mu}}(x|y) Q_{\bar{\mu}}(x,\bar{\mu}'),$$
where,
$$Q_{\bar{\mu}}(x,\bar{\mu}') = R_{\bar{\mu}'}(x) + \gamma \sum\limits_{x'\in X}P_{\bar{\mu}'}(x,x') V_{\bar{\mu}}(x').$$ 
This paper demonstrated how to solve a POMDP without state estimation, by defining a suitable stochastic policy, and showed that it can lead to arbitrarily greater performance than a deterministic policy. Note also, that now the policy space is infinite instead of finite. Nevertheless, we now have a method to create value functions based on observations rather than states. This approach may lead to a computationally fast partially observed multi-agent control algorithm. 






Next, consider the paper \cite{cassandra1996acting}, which develops a number of heuristic MDP based control strategies as well as a few others for a mobile robot navigation problem in an office environment. For the review of this paper I am less interested in the robot architecture and instead focus on the control strategies that were developed as well as the reported results for the agent's effectiveness on the office environments under the different experimental conditions. In these experiments a navigating robot must use the belief framework discussed extensively before to reason about its state.


Briefly, I discuss constructing optimal policies before surveying the heuristics developed. Once a belief state $\pi$ and control $u$ is obtained, there are only $|\mathcal{Y}|$ possible next-stage beliefs, also known as successor beliefs, denoted as $\pi'$. We can define a state transition function for the POMDP, where the state transitions are between belief-states as $\tau(\pi, u, \pi') = \sum\limits_{\{y\in\mathcal{Y} : T(\pi|y,u)=\pi'\}} \sigma(y|u,\pi),$ where $\sigma$ is the normalizing constant in the belief-state update formula. And we can use the notation for belief cost as $\rho(\pi,u) = \sum\limits_{x\in\mathcal{X}}\pi(x)c(x,u)$. Even though we have converted this POMDP into an MDP, the known methods for MDPs only work for finite state and action space, not continuous spaces. The methods that do exist for solving the belief MDP can only solve small scale MDPs with 10 states and 10 observations.

Now, we are in a position to discuss heuristic strategies, which the intractability of the complete POMDP problem motivates. He develops a Maximum Likelihood state estimator approach $\mu_{ML}(\pi) = \mu^*(\arg\max_i \pi(i))$. This paper is worthwhile to review because one could in future research investigations compare the effectiveness of other reduction to MDP methods I have thought of, e.g., the floored mean-state estimator. Also, all of these types of approximations can be applied and studied in a multi-agent control setting, to significantly speed up learning. Many other kinds of reduction to MDP approximations could be compared in an empirical study too that I may have missed in a multi-agent control setting. 


Another idea for future experiments would be to compare these methods with some constraints on periodically updating the belief, and then running VIA from the MDP context. This would be a kind of optimistic value iteration or policy iteration from the POMDP point of view. In the current paper, however the most likely state(MLS) approach selects the control which optimizes the cost-to-go from the state which has the maximum probability in the current belief-state. This takes advantage of the stored optimal policy for the completely known MDP optimal policy $\mu^*$.


The next heuristic is a voting type method. With a belief-state in hand, denote the optimal action probability by $w_u(\pi) = \sum\limits_{x\in\mathcal{X}}\pi(x)\mathbf{1}(\mu^*(x) = u) $, where an indicator function has been used. Then, the most likely optimal control is selected by:
$$\mu_{\textbf{vote}}(\pi) = \arg\max\limits_{u} w_u(\pi).$$ The $Q$-MDP method is reviewed more fully in the next paper \cite{littman1995learning}, but the form of the equation is noted here for completeness of the current paper. $$\mu_{Q_{\textbf{MDP}}}(\pi) = \arg\max\limits_{u} (\sum_x \pi(x)Q(x,u)).$$ This method is more refined than the voting method, as it utilizes the Q function information and the belief-state information. But as discussed before, it is not the best when there is a lot of uncertainties involved in future situations.

Next, and I think this could be useful in RL exploration contexts, particularly for multi-agent control settings, is formulating an entropy definition using the belief-state, and using that to drive the policy construction. That is this method more accurately accounts for different uncertainties in the navigating robot's actions, for example. This method allows agents to explore different controls to obtain more information. Define the entropy of the belief-state $\pi$ as $H(\pi) = -\sum\limits_{x} \pi(x)\log(\pi(x))$, where $\pi(x)\log(\pi(x)) = 0$ if $\pi(x)=0$. The higher this value the more uncertainty, and conversely the lower, the more certain the robot is about it's actions for meeting its objectives.  Next, the author defines the expected entropy as $$EE(u,\pi) = \sum\limits_{\pi'}\tau(\pi,u,\pi')H(\pi').$$ An interesting extension of this idea would be to consider sequences of actions and entropy sequences of some kind to utilize more stochastic methods. This is an open research direction. Also, perhaps an interesting application of this methodology for multi-agent control would be to come up with some time optimization criterion of using expected entropy for some time horizon, and then using $Q$-MDP for the agents in the remaining time of the optimal control problem. 

The next method that is developed uses this entropy definition to define a split policy that takes actions to minimize entropy if the entropy in the belief-state is too high, and conversely take the action that maximizes the voting likelihood if the entropy is below a certain level. The \say{action entropy} policy is $\arg\max\limits_{u}w_u(\pi)$ if $H(w_u(\pi)) < \phi$, where $\phi$ is the threshold. Otherwise, $\arg\min\limits_{u} EE(u,\pi)$. Lastly, we can use an entropy-weighting (EW) approach that balances the two goals just discussed, rather than pursuing either exclusively, weighting each of these aims with a normalized form of entropy. The normalized entropy of a belief-state is defined as $\tilde{H}(\pi) = (\frac{H(\pi)}{H(unif)})^k$, where unif is the uniform pmf, and $k$ is a power that controls the relative weighting of the entropy, thus yielding a useful measure in the compact interval $[0,1]$. The entropy definitions discussed are used to derive new Q-values and a new policy called $\mu_{EW}(\pi) = \arg\max\limits_{u}E[Q(\pi,u)]$. See the paper for more on these definitions. Note the issue for exploration for multi-agent control has not been fully addressed in the literature. 


Now, I present the lessons learned from these methods on the mobile robot navigation problem. These methods were tested in simulation on four different configurations of an office environment with different initial conditions. For these environments, a reward of 1 is given once the agent uses the control declare-goal in the goal state, and a reward of 0 is assigned for every other state-control pair. All episodes were 300 steps long or terminated once the point was earned. All results were averaged over 250 trials. It was found that the MLS method was the most robust across the different settings, illustrating how essential and effective the belief-state info is by itself for determining the optimal policy. The $Q$-MDP method did not do well when it was uncertain in it's starting state. In the final tested scenario, the initial distribution was uniform, so the entropy was maximized. For this scenario, the voting and $Q$-MDP methods did not fare well. In these domains, a reward of 1 is given for performing the action declare-goal in the desired goal state, and 0 for every other state-action pair.

The next paper's \cite{littman1995learning} main idea is to learn a POMDP policy based on the Q-function. $$ \mu_{Q_{MDP}} = \arg\min_{u}\sum_{i}\pi(i)Q^*(i,u) ,$$ where here the Q function is used for cost minimization. In connection with the paper \cite{singh1994learning} on POMDPs and $Q$-learning , here the belief-state is used in coming up with a policy for the POMDP, while before no costly state estimation was required. 

This paper explores approximate methods for finding near optimal POMDP policies, but with the added goal of working within larger state-spaces. The paper reviews several approximation methods for various POMDP problem sizes and shows the different capabilities of each method. The paper explores the theme of promise and practice, in that the optimality of the policy may not have to be sacrificed much in order to solve more challenging problems. RL techniques are developed using insights from the POMDP community. Their algorithm is successfully scaled up to have a hundred states and dozens of observations. Other assumptions made include knowledge of the transition dynamics, and the goal is to find a policy capable of achieving the optimal return. Some concessions are made. The first is that it is fine to have a sub-optimal policy as long as it is close to optimal. Secondly, the problems are all solved and compared from one initial distribution over the states in simulation.



Now, I describe the various solution methods used in this paper and overview their strengths and weaknesses before discussing the successes and failures from the experiments. All methods perform reasonably well in the small domains, but the one dubbed $Q$-MDP performs most consistently. None of the approaches perform well on the two larger problems that are tested. This prompts development of hybrid approaches. The motivation for their algorithms comes from approximating the $Q$-function over the belief-space, and using this to find rewarding controls in the environment. The first method is a truncated form of approximation to the exact VIA methods already discussed by using the finite set of vectors built from learning. The main issue that motivates this and others is that the cardinality of the set of alpha vectors used to approximate the $Q$-function can be exponentially complex from iteration to iteration.

The $Q$-MDP value method is really the core algorithm of this paper, but there are a few other interesting successful algorithms developed. This method uses the $Q$-function computed from the underlying MDP based on the known problem parameters. This approach has been discussed in other papers already surveyed so it is a widely accepted approximation method; this would be a good approximation scheme candidate for the multi-agent control setting. Once we have the $Q$-function from the MDP denoted $Q_{MDP}$, we compute the $Q$-function with belief-state as argument as $Q_\pi(u) = \sum\limits_x \pi(x) Q_{MDP}(x,u)$. This method has the drawback that it will not take exploratory controls because it considers all uncertainty to exist for the current stage. Despite this aspect of the approximation, this method has been effective in practice on moderate to large state-control spaces.


The final algorithm in this section is called Replicated Q-Learning, explored in \cite{chrisman1992reinforcement}. In that paper, they also had to learn the model because they are explicitly RL model-free. The replicated Q-learning algorithm uses one vector $q_u$ to approximate the $Q$-function for every control $u$ as $ Q_\pi(u) = q_u \cdot \pi$. The $q_u$ vector's elements are updated using the formula:$$ q_u(x) = q_u(x) + \alpha\pi(x)[r + \gamma \max\limits_{u'}Q_{\pi'}(u') - q_u(x)]. $$ The update is applied to every $x\in \mathcal{X}$. The learning rate is $\alpha$. The reward is denoted as $r$, the belief-state is denoted by $\pi$, the control is denoted by $u$, and the successor belief-state is denoted by $\pi'$. 


The $Q$-learning rule is applied element wise to every component of $q_u$, multiplied with corresponding weight from the belief-state. Once a sequence of belief-states is obtained the learning rule could be applied, for every pair of belief-states. The update becomes the MDP $Q$-learning rule whenever $\pi(x)=1$ for some $x$. This learning rule applies the $Q$-learning algorithm to each component of $q_u$, weighted by the corresponding component of the belief-state.

The next algorithm developed in the paper is Linear $Q$-learning, which has slightly different learning dynamics from Replicated Q-learning. Instead of adjusting the q-vector evaluated at a state, it adjusts the linear predictor $q_u \cdot \pi$ as follows:


$$q_u(x) = q_u(x) + \alpha \pi(x)[r + \gamma\max\limits_{u'}Q_{\pi'}(u') - q_u \cdot \pi].$$ Similar to the previous algorithm, this algorithm will collapse to the $Q$-learning rule when the belief-state has probability one for only one state. The main limitation of Linear $Q$-learning and the previous algorithm is that they can only learn linear approximations to the true $Q$-function. For the small environments tested, it did not appear to be the case that the policies were poor.

Comparing these algorithms on a range of environments reveals different strengths and weaknesses. In every simulation the agent had the same initial belief-state, and the time horizon was 101 stages long. For the small environments it was the case that the truncated exact VIA had almost identical performance to the exact VIA, and appeared to perform slightly better than the $Q$-MDP algorithm. The $Q$-MDP performed consistently well and was the most time efficient of the group of algorithms. The last two learning algorithms were the slowest at learning.


The next set of experiments looked at environments with 57 states, 21 observations, 4 controls and 89 states, 17 observations, and 5 controls, respectively. To handle these larger POMDPs a hybrid approach to the design of algorithms was considered. Note that none of the algorithms fared well in the larger POMDP setting. This suggested to the authors the possibility of attempting a hybrid solution. One that was tried was computing the $Q$-MDP function and seeding the q-vectors for the learning algorithm. The Linear $Q$-learning algorithm almost doubled its chance of reaching the goal state, and eliminated half the steps required to reach the goal state. Because the navigation environments where so complex, that excluded direct comparison with exact methods.


Nevertheless, this hybrid strategy improved the performance of the algorithms. The authors have called for further improved algorithm design. The main discovery of poor performance of the $Q$-MDP algorithm was that, the agent can end up choosing a stay in place action that forced it in a cycle at a specific belief-state. To conclude this section of the paper I quote, \say{Seeding linear Q-learning using the $Q$-MDP values leads to a promising method of solving larger POMDP's than have been addressed to date. More study is needed to understand the strengths and limitations of this approach.}. More variants of these POMDP approximation methods could be tried on some of the different test-beds mentioned. Particularly, development of POMDP approximation methods for continuous state and control spaces would be interesting, and could be studied under the game theory testbed discussed previously. 


Finally, a call for more advanced representations was proposed by the authors. While piece-wise linear convex functions can approximate any optimal $Q$-function, the linear representation has no such guarantee. Hence different ideas are welcome for solving this problem of good representation of the optimal Q-function that is easy to compute and time-efficient. To me, this is a big open area of research, and the first experiment would be to try a new algorithm called Quadratic $Q$-learning, that uses a quadratic basis, possibly with linear and bias terms as well; other sets of basis functions could be compared on the performance of various sized POMDP learning tasks. In closing, the authors proposed a PWLC Q-learning algorithm that maintains a set of vectors for every possible control and uses an update rule based on competition between the vectors. It goes as follows. Essentially, the belief-state dotted with the $q$-vector that produces the largest value is used for updating. The updating rule used is the linear Q-learning rule. The hope is that this approximation allows a better coverage of the state-space then is possible with a single vector representation for the linear function. They tested this algorithm with three q-vectors per action. When a new belief-state, value pair arrives, the $q$ -vector with the largest dot product is used for updating.

\section{Conclusion}
\label{conclusion}

In conclusion we have studied POMDPs and the main ways of developing approximate control strategies for them; however, there are many more possible modelling variations that need to be researched for the POMDP and its approximation. Now I list a few of the possible research directions from here for approximate methods for POMDPs: exploration strategies in POMDPs, applying these approximate POMDP strategies in multi-agent control settings and developing guarantees for such multi-agent systems. Information theory should help out or be a starting point for developing useful exploration strategies and may lead to even better approximate control strategies. For the approximate methods for POMDPs covered in this survey, those methods could be extended to multi-agent Markov decision processes (MMDPs), stochastic games (SGs), or Decentralized POMDPs (Dec-POMDPs). 


\bibliographystyle{IEEEtran}
\bibliography{references}
\end{document}